\documentclass[pdflatex,sn-mathphys-num]{sn-jnl}

\usepackage{graphicx}%
\usepackage{multirow}%
\usepackage{amsmath,amssymb,amsfonts}%
\usepackage{amsthm}%
\usepackage{mathrsfs}%
\usepackage[title]{appendix}%
\usepackage{xcolor}%
\usepackage{textcomp}%
\usepackage{manyfoot}%
\usepackage{booktabs}%
\usepackage{algorithm}%
\usepackage{algorithmicx}%
\usepackage{algpseudocode}%
\usepackage{listings}%
\usepackage{subfigure}
\usepackage{bbm}

\newcommand{\D}{\mathrm{d}}

\DeclareMathOperator*{\argmin}{arg\,min}

\begin{document}

\title[Article Title]{{\bf Scalable AI Uncertainty Quantification via Generalized Laplace Active Subspaces}}


\author[1,3]{\fnm{Wouter N.} \sur{Edeling}}\email{edeling@cwi.nl}

\author*[2,4]{\fnm{Peter V.} \sur{Coveney}}\email{p.v.coveney@ucl.ac.uk}


\affil*[1]{\orgdiv{Scientific Computing Group}, \orgname{CWI}, \orgaddress{\street{Science park 123}, \city{Amsterdam}, \postcode{1098XG}, \country{the Netherlands}}}

\affil[2]{\orgdiv{Centre for Computational Science}, \orgname{University College London}, \orgaddress{\street{20 Gordon Street}, \city{London}, \postcode{WC1H 0AJ}, \country{UK}}}

\affil[3]{\orgdiv{Faculty of Electrical Engineering, Mathematics and Computer Science}, \orgname{University of Twente}, \orgaddress{\street{Hallenweg 15}, \city{Enschede}, \postcode{7500AE}, \country{the Netherlands}}}

\affil[4]{\orgdiv{Advanced Research Computing Centre}, \orgname{University College London}, \orgaddress{\street{ Gordon Street}, \city{London}, \postcode{WC1E 6BT}, \country{UK}}}


\abstract{Reliable uncertainty quantification (UQ) is essential for deploying neural networks in scientific and high-stakes applications, but full Bayesian inference over the network parameters is computationally infeasible. We propose a low-rank generalized Laplace approximation for neural-network UQ based on a small number of data-informed curvature directions. Starting from a generalized Bayesian posterior defined through an empirical loss, we construct a local Gaussian approximation around a pretrained set of weights in this active curvature subspace. The posterior variances in the retained subspace are available in closed form, and the prior variance is calibrated by an empirical Bayes procedure. The generalized Bayesian formulation allows us to compare two posterior scalings: the standard Bayesian scaling associated with the summed negative log likelihood, and a mean-loss scaling in which the empirical loss is normalized by the number of data. A central finding is that the standard scaling induces a data-size dependent contraction of the posterior variance in the leading active directions. In regression problems, this can force the low-rank framework to retain additional weak-curvature directions in order to achieve nominal coverage of calibration data. When posterior samples are propagated through the non-linear network, these additional directions can degrade the coherence of the predictive intervals and shift the posterior predictive mean away from the pretrained model. In contrast, the generalized mean-loss scaling yields a more stable, lower dimensional active subspace and produces calibrated, coherent predictive confidence intervals. These results indicate that generalized Laplace active subspaces provide a practical and scalable route to calibrated uncertainty quantification in neural networks. We demonstrate the method on two regression problems, and a classification case involving an autoregressive molecular language model.}

\keywords{uncertainty quantification, neural networks, Laplace posterior, active subspaces}



\maketitle

\section{Introduction}

One of the biggest open questions concerning artificial intelligence (AI) is the reliability and trustworthiness of the predictions which emanate from its output. After training, AI software operating in inference mode can spew out predictions concerning the behaviour of a new system in the same class as  those it has been trained on in virtually instantaneous fashion. Such predictions can have enormous value in many real world scenarios. In some situations, qualitative classifications suffice while others require only a modicum of accuracy to be beneficial; still others require accurate and precise predictions or they are not actionable.  This last scenario is the one which is most often necessary in scientific and engineering contexts.

To use the results of a calculation inside a scientific procedure, one generally needs to know that the predictions are reliable within defined and suitably limited bounds of uncertainty.  Actionable predictions must therefore not only be timely; they need to be of a sufficient level of accuracy and/or precision to be used with confidence. The problem with AI methods in general is that what they output often appears plausible and may pass muster as such for some applications; but in more quantitative scientific and mathematical domains, plausibility is not sufficient: we need to be clear about the level of certainty of its predictions. The key assessments of validation, verification and uncertainty quantification are regularly visited on physics-based models, e.g.\ for weather and climate prediction \cite{palmer2000predicting, murphy2004quantification}, turbulence modelling \cite{cheung2011bayesian, gorle2013framework}, tsunami hazards \cite{liu2017dimension, selva2016quantification} or biological applications \cite{avitabile2026neural}. In these instances there is an exactly defined mathematical model based on clearly defined equations that typically include a variety of parameters, whereas such systematic analysis are significantly more challenging for AI systems. 

One immediately noticeable difference between AI and physics-based codes resides in the number of parameters which arise in the two cases. Whereas physics-based codes typically contain around five parameters \cite{cheung2011bayesian}, possibly several hundreds \cite{edeling2024global} (extending to a few thousand in some large scale codes), each parameter has a well-defined physical meaning pertaining to the underlying model. In AI software there are many orders of magnitude more, none of which has a physical interpretation, as they are simply free parameters. The impressive fitting of input-output data can be readily performed with so many free parameters in play, particularly when one works with held-out data sets where all manner of optimisations can be conducted. The approach lends itself to the making of overconfident predictions, which can and often do unravel when the trained system is applied to data which it has not previously encountered.  We are interested in the uncertainty of such predictions in terms of the uncertainty in the parameter values of the trained model; unlike a physics-based approach, lacking any physical meaning, these parameters cannot be meaningfully adjusted based on insights into the system which can be gleaned from  mathematical models we have created.

The purpose of the present paper is to demonstrate that, despite the lack of clarity as to the meaning of the parameters in an AI system, it is possible to obtain predictive uncertainty estimates. A natural framework for quantifying parameter uncertainty is Bayesian inference \cite{gelman2013bayesian}. In principle, one would place a prior distribution on the neural-network weights, update this prior by conditioning on data, and propagate the resulting posterior distribution through the network to obtain predictive uncertainty. In practice, since the parameter dimension is extremely large, standard sampling strategies are computationally infeasible. Scalable approximations are therefore unavoidable.

Scalable Bayesian neural-network posterior approximations include variational  approaches such as Bayes by Backprop~\cite{blundell2015weight}, which learns a parametric approximation to the posterior distribution over the network weights, typically using a Gaussian variational family. The resulting training objective is an evidence lower bound, consisting of  an expected data-fit term and a Kullback-Leibler (KL) divergence to the prior. 
In practice, the balance between these two terms can be delicate, and may require  specialized training strategies such as KL annealing or cyclical schedules \cite{fu2019cyclical}.

By contrast, the Laplace approximation~\cite{mackay1992practical} is a post-hoc posterior approximation.  It also models the parameter posterior by a Gaussian distribution, but this Gaussian is centred at a pre-trained set of weights, and its covariance is obtained from the local curvature of the log posterior density rather than by optimizing a variational distribution during training. While the posterior is now conceptually simple, the computational bottleneck shifts to the covariance matrix of the Laplace approximation, as it is too large to form explicitly for many networks. In this work we circumvent this bottleneck via a generalized-Bayes low-rank Laplace approximation. The posterior is restricted to a low-dimensional active subspace spanned by the dominant Gauss-Newton curvature directions of the empirical mean loss. These directions are computed using matrix-free curvature-vector products, so that the full covariance matrix is never formed. Although active subspaces are originally derived from the dominant directions of an uncentered covariance matrix of the gradient of a (physics-based) model \cite{constantine2014active}, the overarching idea is similar, namely to find a linear projection to a low-dimensional subspace relevant for a quantity of interest. For the original active subspace approach this quantity was the model output subject to a prior distribution, in this work it is the posterior predictive uncertainty of a neural network. Within the subspace spanned by these dominant active directions, the posterior variances are available in closed form, while the prior variance is calibrated using a generalized-evidence criterion. The active subspace dimension is selected by empirical coverage on held-out calibration data, yielding predictive intervals that are both calibrated and sharp.

We derive our low-rank Laplace posterior within the framework of generalized Bayesian inference, where the data is connected to the parameters via a loss function, rather than a traditional likelihood function \cite{bissiri2016general}. In particular we contrast summed with mean loss functions, where the latter is scaled by $1/N$, $N$ being the number of data records in the training set. The use of the summed loss equates to the standard Bayesian likelihood function. A central observation of the paper is that the scaling of the generalized posterior is not merely a technical detail. The standard Bayesian scaling leads to an increasingly concentrated posterior variance with the number of training data. To avoid such overconfident predictions, a low-rank approximation should then retain many weak-curvature directions with relatively high posterior variance. This increases the computational cost, and reduces the statistical coherence of the predictive output distribution. In other words, when samples from a local Gaussian weight posterior are propagated through the nonlinear network, these weak directions can shift the predictive mean or distort the resulting uncertainty intervals. Existing methods \cite{immer2021improving, faller2026low} deal with this issue by replacing the full network with its linear expansion.  By contrast, our generalized Bayesian mean-loss scaling avoids the purely data-size-driven contraction, results in a much lower-dimensional active subspace and can retain the full non-linear network. Numerical studies demonstrate that the proposed approach yields coherent predictive intervals with only a small number of active directions.


\section{Results\label{sec:res}}


Let $f(x;\theta)\in\mathbb{R}^M$ represent our Quantity of Interest (QoI), namely the output of a pretrained neural network. Here $x$ are the inputs to the neural network, whereas $\theta\in\mathbb{R}^D$ is the collection of all connection weights. Furthermore, let $(x, y)$ denote all supervised training data, and specific independent identically-distributed training samples are indexed as $(x_n, y_n)$, $n=1,\cdots,N$, with $y_n\in\mathbb{R}^M$. It is our goal to perform scalable uncertainty quantification of $f$ by propagating a low-rank Bayesian posterior distribution on $\theta$ through the network.

Before we describe our results, we give a brief overview of our UQ framework, called generalized Laplace active subspaces. Full details are found in the Methods section; the PyTorch code and data to reproduce the results below are available through \cite{edeling2026glas}.

\subsection{Generalized Laplace active subspaces}

Two choices define our Bayesian posterior. First, instead of seeking a posterior distribution $p(\theta\mid y)$ in the original $D$-dimensional coordinate system of the connection weights, we seek a low-rank ($d\ll D$) coordinate system that is most relevant for the posterior predictive uncertainty. Secondly, we do not employ the standard Bayesian belief update $p(\theta\mid y)\propto p(y\mid\theta)p(\theta)$, which uses the likelihood function $p(y\mid\theta)$ to connect the parameters $\theta$ to the observed data $y$. For independent Gaussian observations, the corresponding negative log-likelihood is proportional to a summed loss. In contrast, neural networks are typically trained using an average loss $\mathcal{L}_N$, such as the mean squared error, which includes an additional factor $1/N$. To obtain a Bayesian-style update that is consistent with the average loss used in practice, we adopt the generalized Bayesian framework of \cite{bissiri2016general} instead, which gives rise to the following posterior:
\begin{align}
    \pi_\beta(\theta\mid y) \propto \exp\!\left[-\beta \mathcal L_N(\theta)\right]p(\theta),
\end{align}
\noindent
see Section \ref{sec:gen_bayes} for more details. The so-called `temperature' parameter $\beta$ controls the contribution of the loss relative to the prior. This formulation includes the standard Bayesian posterior as a special case, obtained by the appropriate choice of $\beta$, while also allowing alternative scalings such as the mean-loss scaling considered in this work. It therefore provides a convenient framework for studying how posterior scaling (i.e.\ the choice of $\beta$) affects the dimension and structure of the resulting low-rank posterior approximation.

This posterior approximation is obtained by constructing a local Laplace approximation centered at the pre-trained weights $\theta_0\in\mathbb{R}^D$. With a commonly-assumed isotropic Gaussian prior $p(\theta)=\mathcal{N}(0,\sigma^2_0I_D)$, this results in a Gaussian posterior distribution $\pi_\beta(\theta\mid y) = \mathcal{N}(\theta_0, \Sigma)$, with 
\begin{align}
\Sigma = \left(\beta H_L + \frac{1}{\sigma_0^2}I_D\right)^{-1}.
\end{align}
\noindent
Here, $H_{L}$ is defined to be the Hessian (curvature) of the mean loss: $H_L:=\nabla^2_{\theta}\mathcal{L}_N(\theta_0)\in\mathbb{R}^{D\times D}$.  For large neural networks this matrix cannot be formed or inverted explicitly, and it may not be positive semi-definite.  We therefore replace the full curvature by a low-rank approximation in the dominant eigenspace of the generalized Gauss-Newton matrix $G$ (defined by \eqref{eq:G}), $G$ being a positive semi-definite approximation to $H_L$. The $d$-dimensional dominant $G$ eigenspace with $\lambda_1\geq\lambda_2\geq\cdots\geq\lambda_d$;
\begin{align}
    G p_j = \lambda_j p_j,
    \quad
    j=1,\ldots,d,   
\end{align}
\noindent
defines the projection matrix $P = [p_1,\ldots,p_d]\in\mathbb{R}^{D\times d}$, used to obtain the posterior perturbations on top of $\theta_0$ via
\begin{align}
    \theta = \theta_0 + P z,
    \quad
    z\in\mathbb{R}^d.
    \label{eq:post_theta}
\end{align}
In this way uncertainty is represented only via the dominant curvature-informed directions, along which the data was able to inform the parameters well. In the coordinate system of this subspace the Laplace posterior has diagonal covariance, with
\begin{align}
    \sigma_j^2 =
    \frac{1}{\beta\lambda_j+ \sigma^{-2}_0}.
\end{align}
Thus, once the leading eigenpairs $(\lambda_j, p_j)$ of $G$ are available, the posterior variances $\sigma_j^2$ in the retained active directions are obtained in closed form. The eigenpairs are computed using automatic differentiation, matrix-free vector products $Gv$ and the Lanczos algorithm (see section \ref{sec:matrix_free}). This avoids explicitly forming $G$, and makes the construction applicable to networks with a large number of parameters. The resulting approximation is therefore both local and low-rank: local because it describes uncertainty around a single trained solution \(\theta_0\), and low-rank because only a small number of active directions are retained.

The prior variance $\sigma_0^2$ sets the overall scale of the posterior, and can be calibrated by an empirical-Bayes criterion within the active subspace $w_0=P^T\theta_0\in\mathbb{R}^d$. The resulting prior variance is found by solving
\begin{align}
\lVert w_0\rVert^2_2 = \sum_{i=1}^d \frac{1}{\alpha} - \frac{1}{\beta\lambda_i + \alpha} = \sum_{i=1}^d \sigma^2_0 - \sigma^2_i,
\label{eq:sigma_0_2}
\end{align}
\noindent
for $\alpha = \sigma^{-2}_0$. Note that this relation chooses the prior scale such that the norm of the projected parameter vector is consistent with the amount of variance reduction induced by the data in the active subspace.


The posterior predictive distribution is then approximated by sampling \eqref{eq:post_theta} and propagating these samples through the neural network $f(x;\theta)$. This sampling strategy is denoted as sampled Laplace. However, the non-linear nature of the network may distort the UQ estimates. So-called linearized Laplace methods \cite{immer2021improving}, which replace the full network with its linear expansion (see \eqref{eq:f_lin}), are a means to counteract this. We will contrast these two sampling methods below.

\subsection{Regression with known observational noise}

We start with the same scalar ($M=1$) regression function with Gaussian noise as in \cite{jantre2024learning}, namely; 
\begin{align}
 y(x) = \sin(4\pi x) + \sin(7\pi x) + \epsilon, \quad \epsilon\sim\mathcal{N}(0, \sigma^2), \quad x\in[0,1],
\label{eq:f_analytic}
\end{align}
\noindent
using a known noise of $\sigma=0.1$. 

\begin{figure}[htp]
\centering
\includegraphics[scale=0.6]{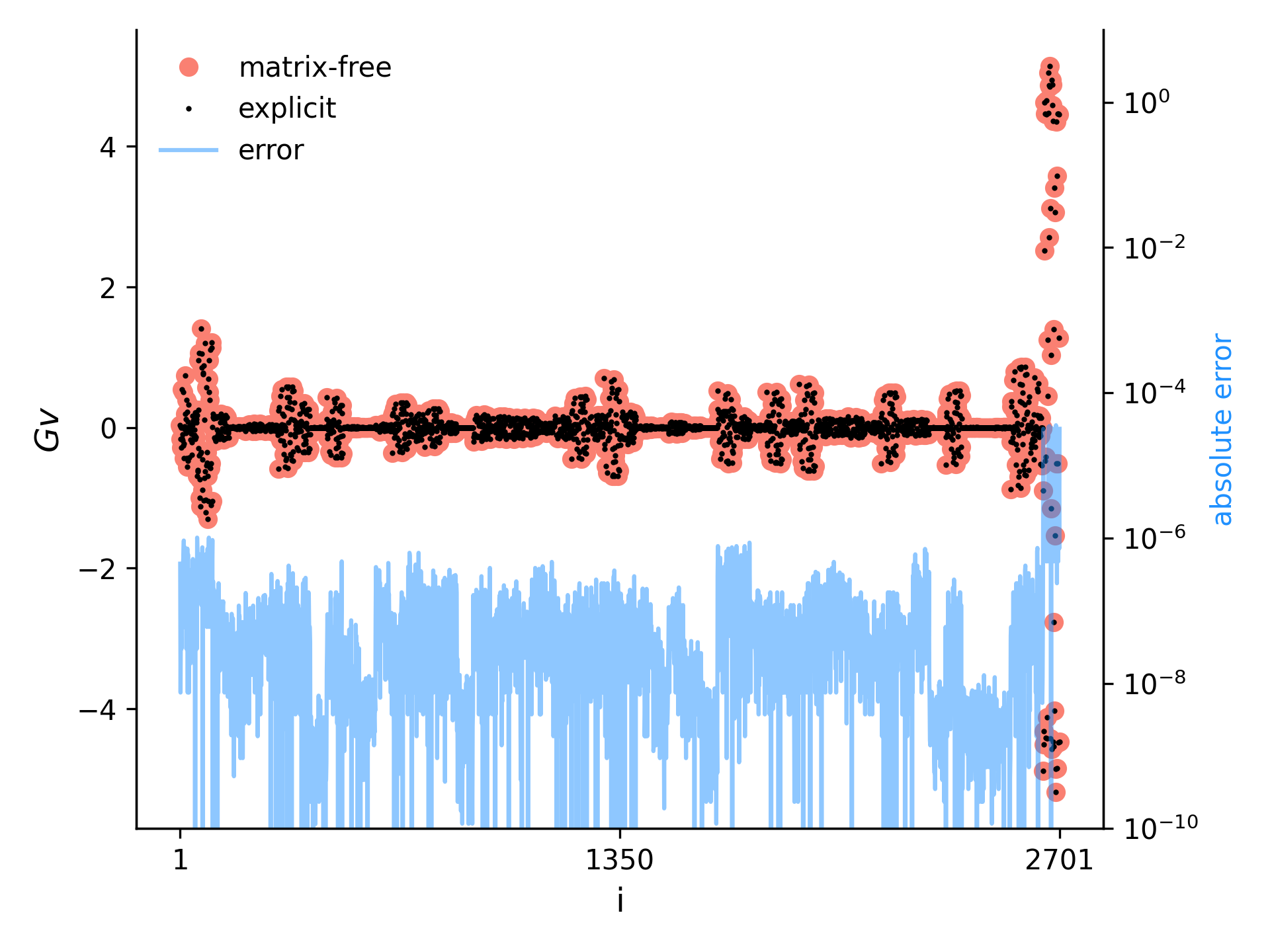}
\caption{The matrix-vector product $Gv\in\mathbb{R}^D$ with $D=2701$, computed using the matrix-free method outlined in Section \ref{sec:matrix_free}, and by explicitly forming matrix $G$ via \eqref{eq:G}. The absolute error $|Gv - (Gv)_{explicit}|$ is plotted on the right vertical axis. \label{fig:Gv}}
\end{figure}

\begin{figure}[htp]
\centering
\subfigure[Eigenvalue fit and errors.]
{\includegraphics[scale=0.5]{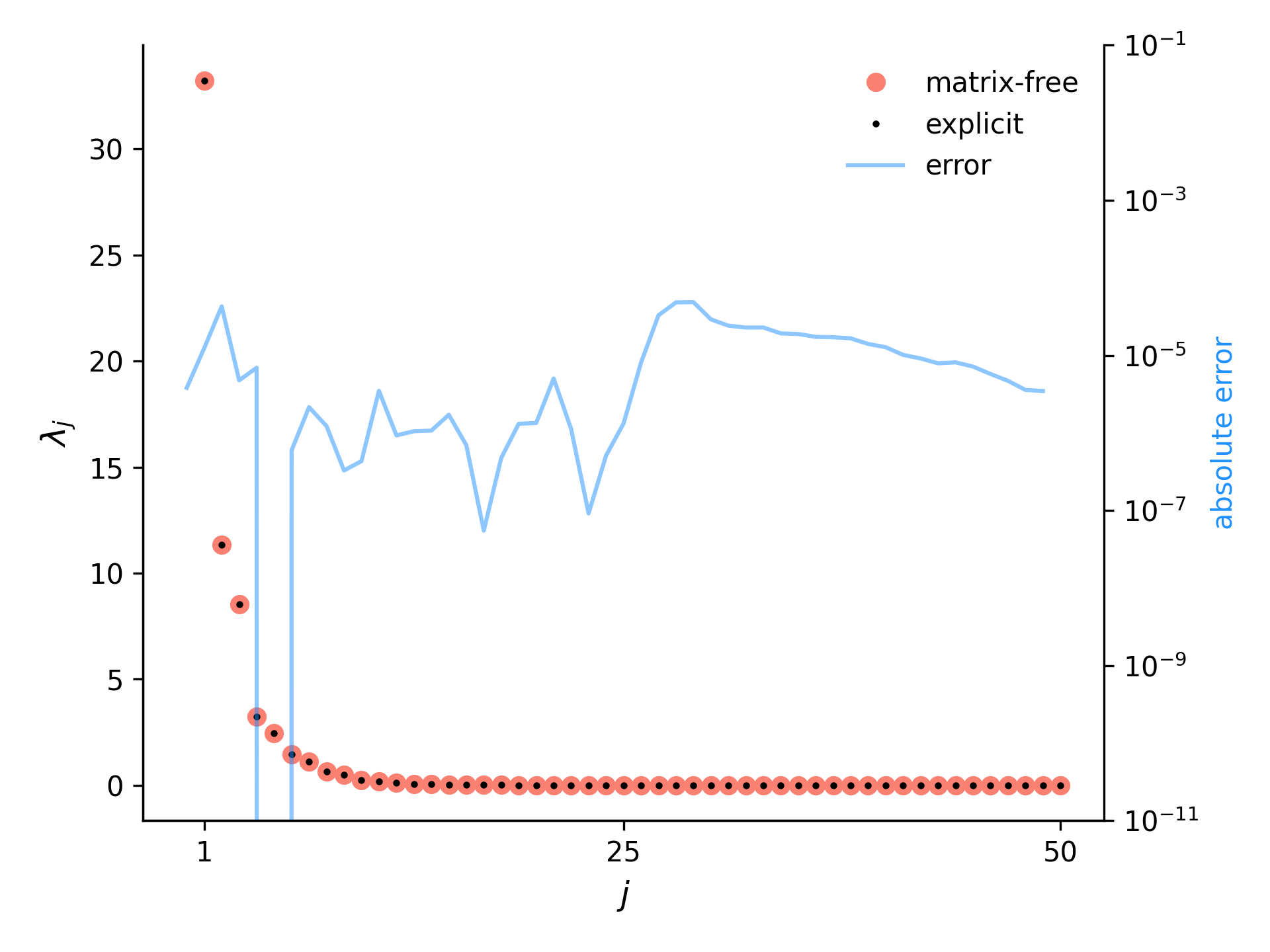}}
\subfigure[Eigenvector fit and errors]
{\includegraphics[scale=0.5]{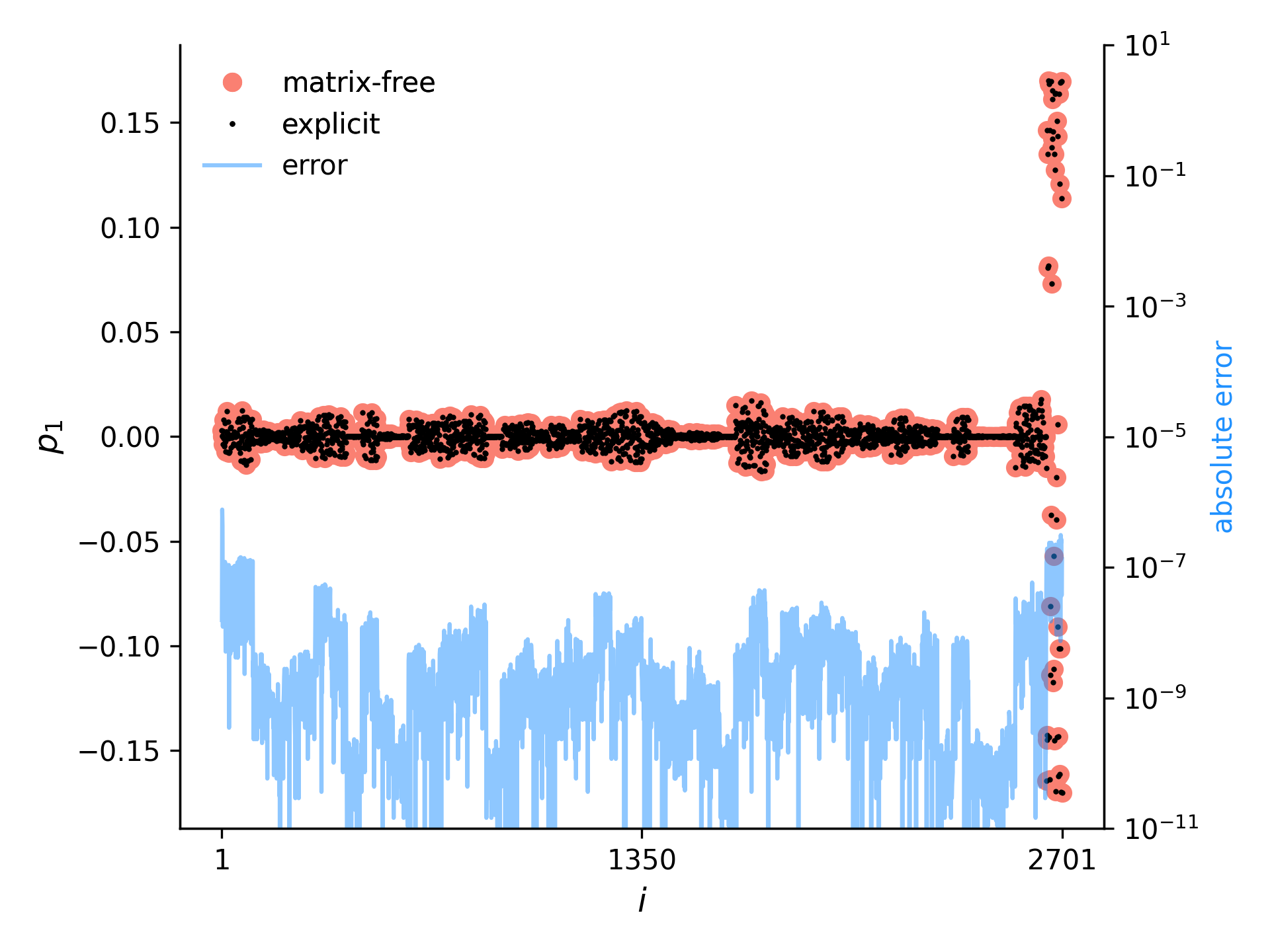}}
\caption{The $d$ leading eigenvalues $\lambda_j$ and leading eigenvector $p_1$ of $G$, computed using the matrix-free method outlined in Section \ref{sec:matrix_free}, and by explicitly forming matrix $G$ via \eqref{eq:G}. The absolute errors are plotted on the right vertical axis. \label{fig:eigvals_p1}}
\end{figure}

{\bf Implementation verification.} This example is used to verify the implementation, in particular the matrix-free computation of $Gv$ and the dominant eigenpairs of $G$. We train a fully-connected feed-forward neural network, using back propagation with
Stochastic Gradient Descent and the Adam optimizer \cite{kingma2014adam, aggarwal2018neural}. The learning rate was set at 0.001, and we used two hidden layers at 50 neurons each. The dimension of the weight space is $D=2701$, small enough to explicitly form the full $G$ matrix \eqref{eq:G}. The matrix-free $Gv$ vector is contrasted against the corresponding explicit matrix-vector product in Figure \ref{fig:Gv}. The absolute error $|Gv - (Gv)_{explicit}|$ is $\mathcal{O}(10^{-8})$ in general (using the \texttt{torch.float32} data type). A similar verification for the matrix-free eigenpairs is displayed in Figure \ref{fig:eigvals_p1}. 

{\bf Standard vs generalized scaling.} Next we examine the scaling of the output confidence intervals (CIs) with the temperature parameter $\beta$, with the CIs computed using the sampled Laplace strategy. As described in Section \ref{sec:gen_bayes}, the value $\beta = N\sigma^{-2}$ corresponds to the standard Bayesian setting, whereas with $\beta = \sigma^{-2}$ we employ the generalized Bayesian scaling where the task of standard likelihood function is taken up by the mean-squared loss. For both posterior scalings we determine the active subspace dimension $d$ as described in Section \ref{sec:d}. Essentially, we increase $d\in\mathbb{N}$, starting from $d=1$ and solve \eqref{eq:sigma_0_2} for $\sigma^2_0$ every iteration, until our computed 90\% CIs achieve empirical coverage of (at least) $90$\% of held-out calibration data. Our goal here is to compare this minimal required subspace dimension $d$ for both posterior scalings.

\begin{figure}
\centering
\subfigure[$\beta = \sigma^{-2}$, $N=50$, $d=2$ \label{fig:beta_a}]
{\includegraphics[scale=0.5]{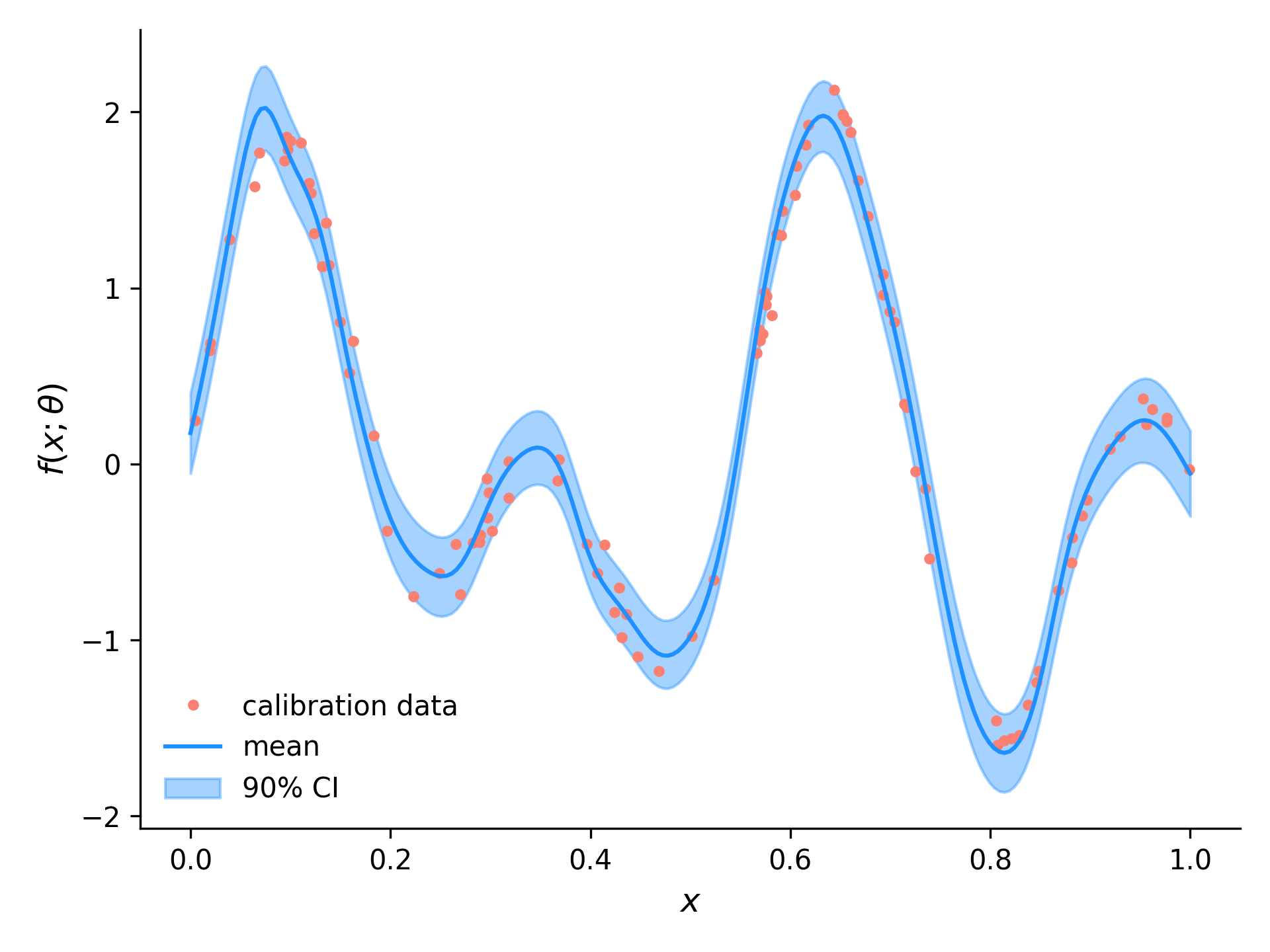}}
\subfigure[$\beta = N\sigma^{-2}$, $N=50$, $d=31$ \label{fig:beta_b}]
{\includegraphics[scale=0.5]{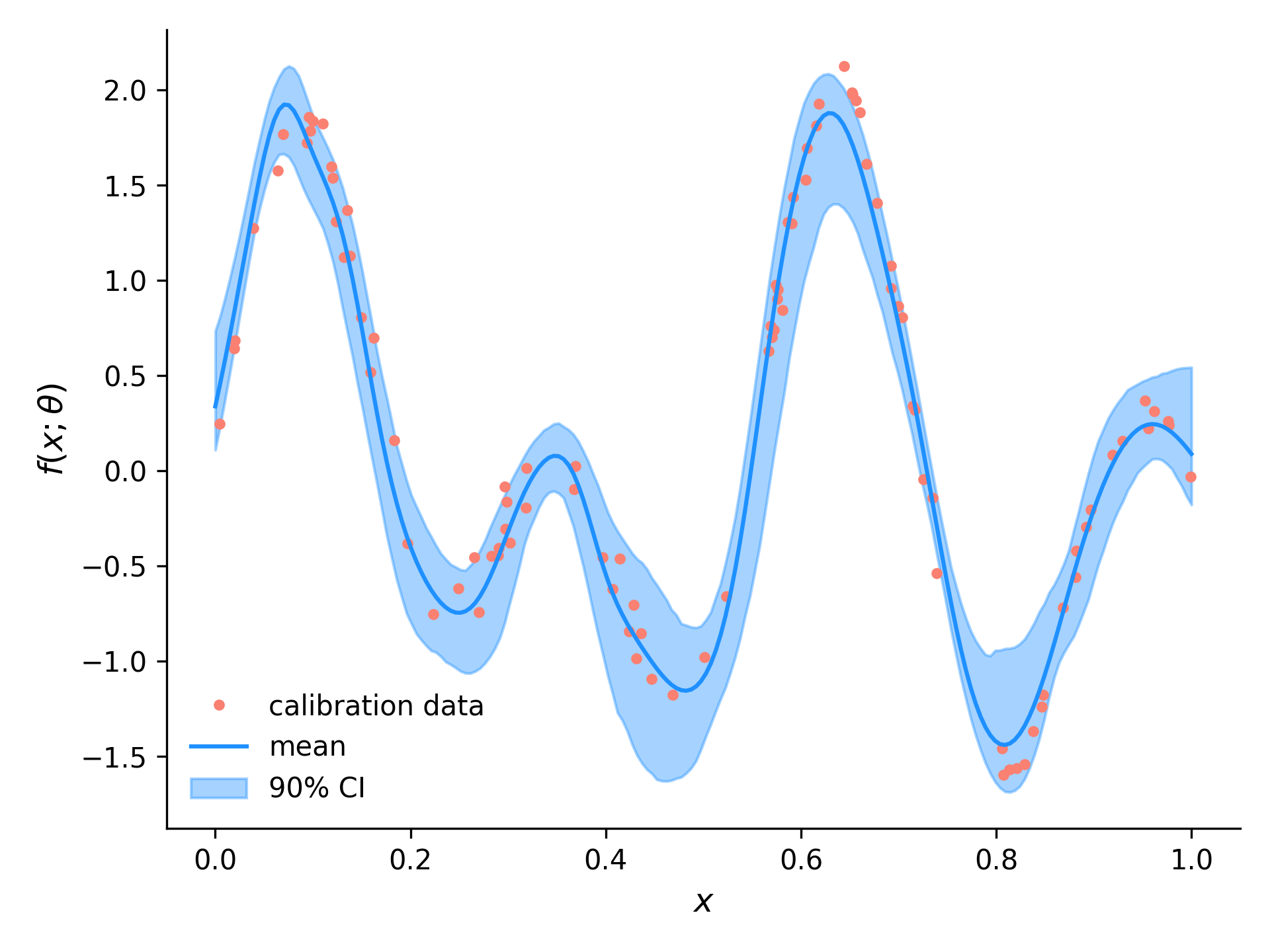}}
\caption{90\% confidence intervals for the, a) generalized Bayesian scaling, and b) standard Bayesian scaling. Both CIs were generated by drawing 1000 posterior samples, and propagating these through the network. \label{fig:beta}}
\end{figure}

Figure \ref{fig:beta} displays the result. The generalized Bayesian posterior only required $d=2$ active subspace directions to achieve coverage, and produces coherent CIs. Conversely, the standard Bayesian scaling of $\beta$ required $d=31$ directions and yields slightly wider, less coherent intervals. 

\begin{figure}
\centering
\includegraphics[scale=0.5]{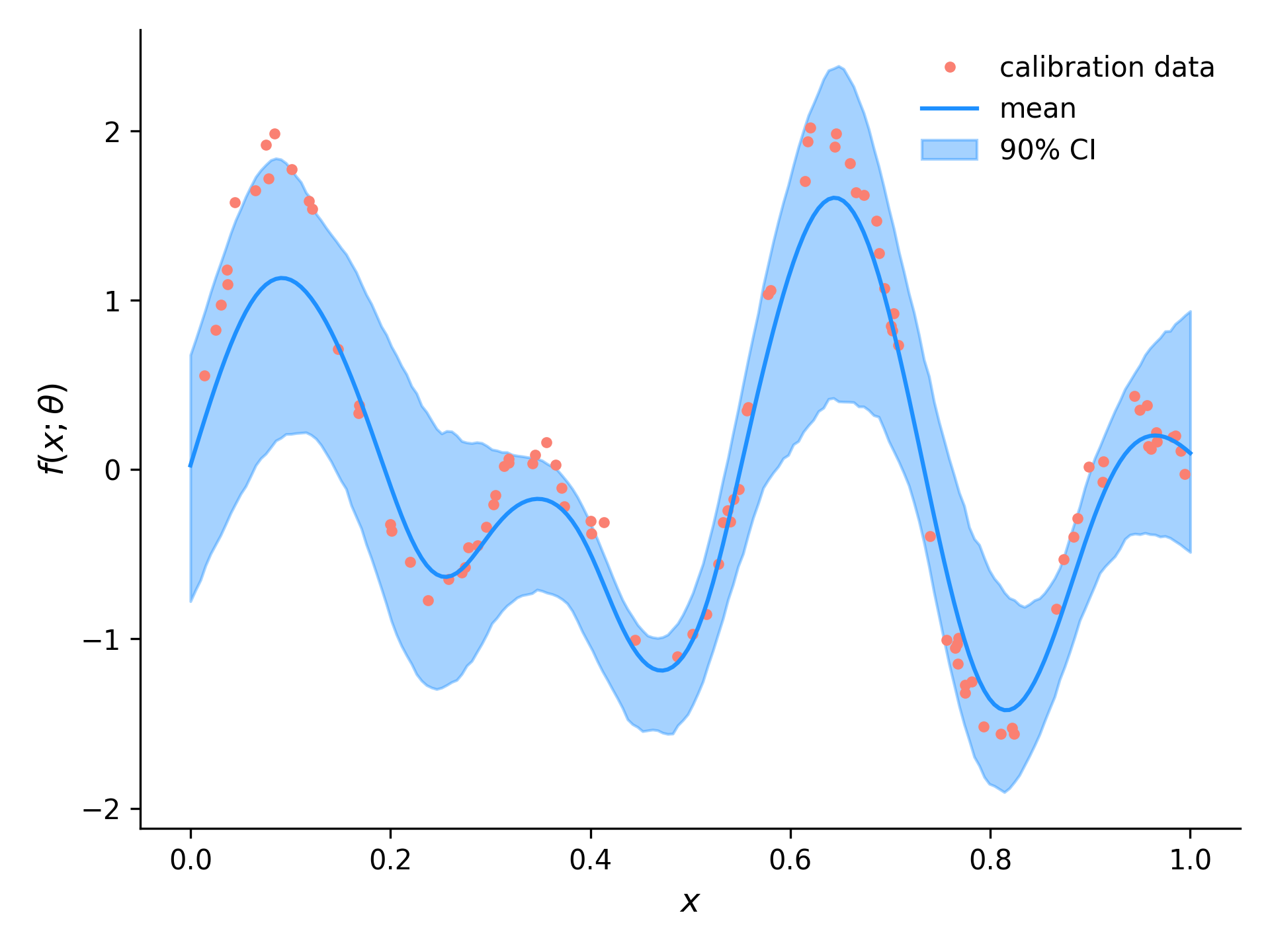}
\caption{The results for the standard Bayesian scaling $\beta=N\sigma^{-2}$ with $N = 1000$ and $d=43$. The CIs were generated by drawing 1000 posterior samples, and propagating these through the network. \label{fig:N1000}}
\end{figure}

The preceding results were obtained with $N = 50$ training data points. We now compare the standard and generalized scaling using $N=1000$. The results from the generalized Bayesian scaling (not shown) are essentially unchanged by the increase in training data, as it still yields a 2D active subspace that generates consistent CIs. We therefore focus on the 90\% CI results of the standard Bayesian scaling, as displayed in Figure \ref{fig:N1000}. Clearly, the quality of the mean prediction and the CIs significantly degraded. Furthermore, instead of $d=31$, we now required 43 curvature directions to achieve coverage.

\begin{figure}
    \centering
    \subfigure[$N=100$, $d=50$, 75\% coverage \label{f_lin_a}]
    {\includegraphics[scale=0.45]{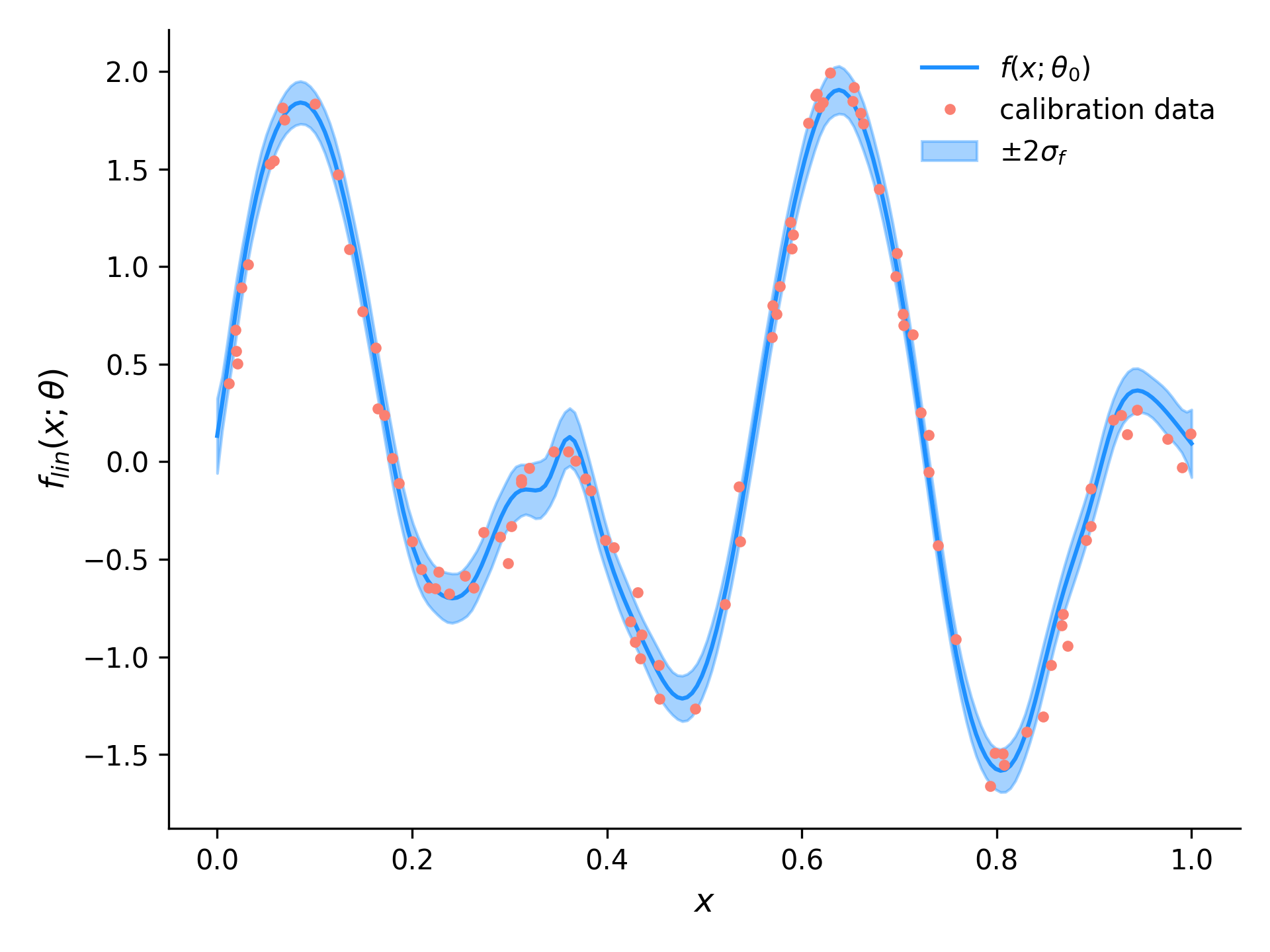}}
    \subfigure[$N=100$, $d=50$, 40\% coverage]
    {\includegraphics[scale=0.45]{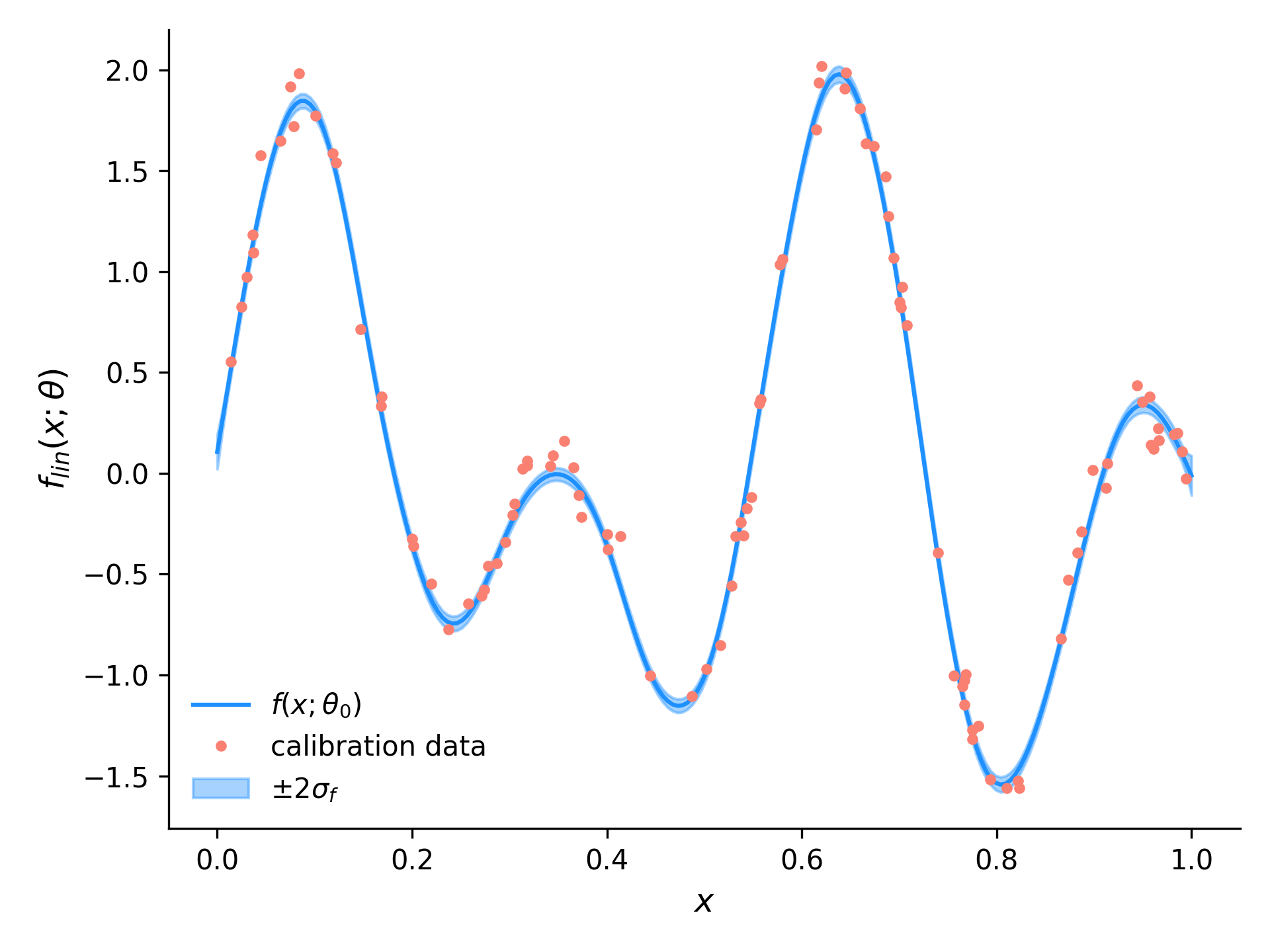} \label{f_lin_b}}
    \caption{The results for the standard Bayesian scaling $\beta=N\sigma^{-2}$ combined with the linearized Laplace predictive posterior distribution, computed using $f(x;\theta_0) \pm 2 \sigma_f$. \label{fig:f_lin}}
\end{figure}

{\bf Sampled vs linear Laplace.} The underfitted results of Figure \ref{fig:N1000} are a consequence of the of propagating the Laplace posterior through the non-linear network, see also \cite{immer2021improving}. Instead, the posterior predictive results obtained from linearized Laplace $f_{lin}$, i.e.\ from \eqref{eq:f_lin}, are shown in Figure \ref{fig:f_lin} for $N=100$ and $N=1000$ for the standard scaling. Due to the linearity of the predictive model, its distribution is Gaussian as well, with a standard deviation $\sigma_f$ given by \eqref{eq:sigma_f}. We therefore perform the coverage calibration using a 95\% confidence interval constructed from $f(x;\theta_0) \pm 2 \sigma_f$, which avoids Monte Carlo sampling. We further imposed a maximum subspace dimension of $d_{max}=50$.

As expected, the posterior predictive mean is improved relative to Figure \ref{fig:N1000}, since the linearized predictive distribution is centered by construction at the expansion-point prediction $f(x;\theta_0)$. The confidence intervals are substantially narrower, confirming that the overly wide intervals observed for non-linear posterior sampling are largely caused by propagating weight samples through the non-linear network. However, the linearized predictive distribution is under resolved. For both $N=100$ and $N=1000$, the calibration procedure reaches the maximum allowed dimension of 50, covering only approximately 75\% and 40\% of the calibration data, respectively, with nominal 95\% confidence intervals. 

\subsection{Regression with unknown observational noise}

The following test case involves the Concrete Compressive Strength dataset from the UCI machine learning repository, see \cite{concrete_compressive_strength_165} for details. This is a regression test case containing $N=1030$ samples and 8 input features describing the concrete mixture, and a scalar output corresponding to compressive strength. We train the same architecture as the preceding test case, except that now it contains approximately $D=10^6$ connection weights. This case therefore assesses the scalability of the proposed matrix-free low-rank generalized Laplace approximation. 

Since the observation noise is unknown, we estimate the noise $\sigma$ via residuals:
\begin{align*}
\sigma^2\approx \frac{1}{N}\sum_{n=1}^{N} \lVert y_n - f(x_n; \theta_0)\rVert_2^2,
\end{align*}
which is approximately 0.07 for this test case, and we use this value to specify $\beta$ as before. We set the number of training data $N$ to be 90\% of the 1030 available data points. The remainder ($N_{cal}$) is used to calibrate $d$. We execute the calibration procedure until our target coverage of 90\% is met, or until a preset maximum active subspace dimension $d_{max} = 50$ is exceeded. 

\begin{figure}
\centering
\subfigure[Sampled Laplace posterior prediction, 90\% nominal coverage. \label{fig:coverage_a}]
{\includegraphics[scale=0.38]{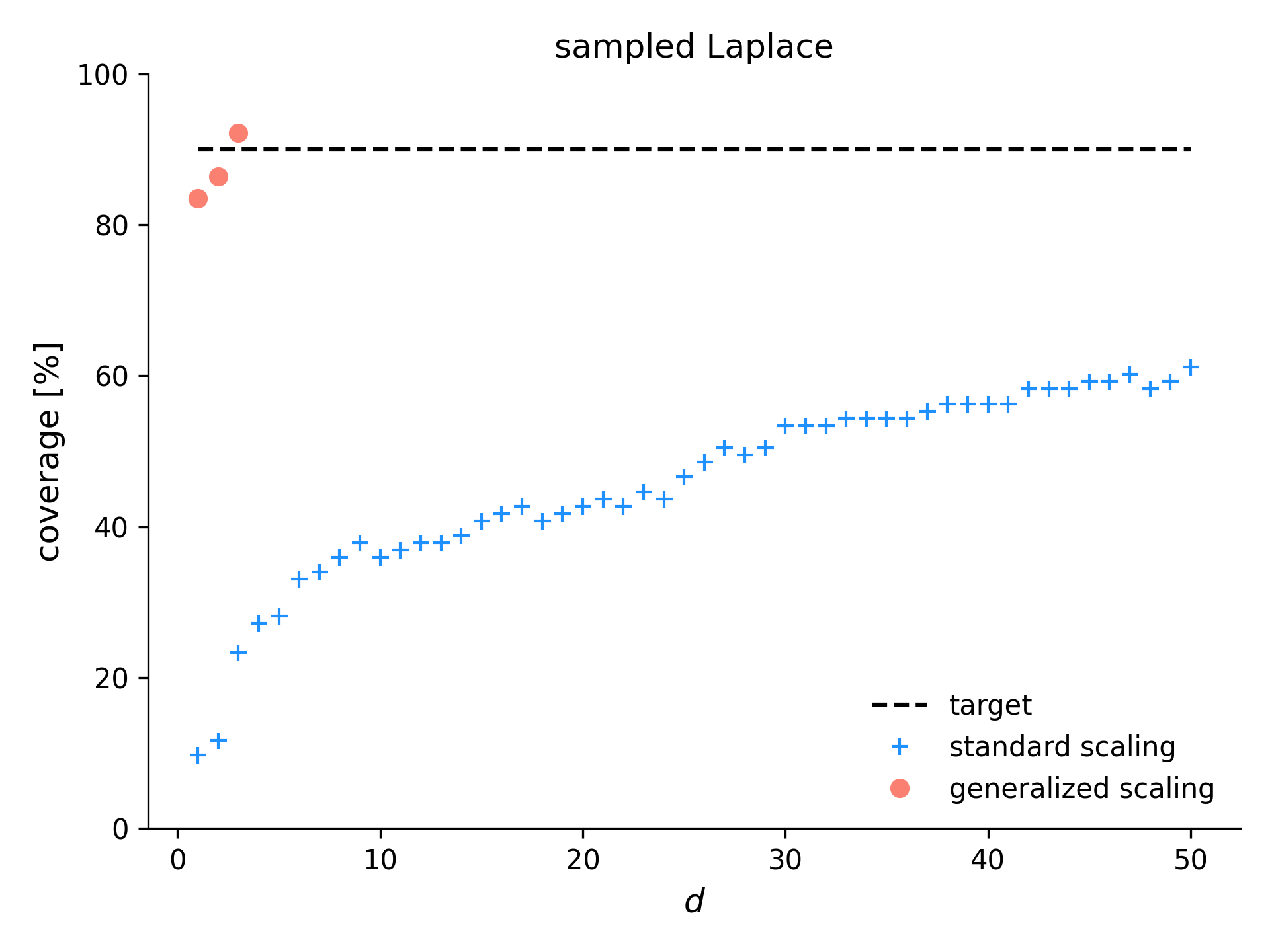}}
\subfigure[Linear Laplace posterior prediction, 95\% nominal coverage. \label{fig:coverage_b}]
{\includegraphics[scale=0.38]{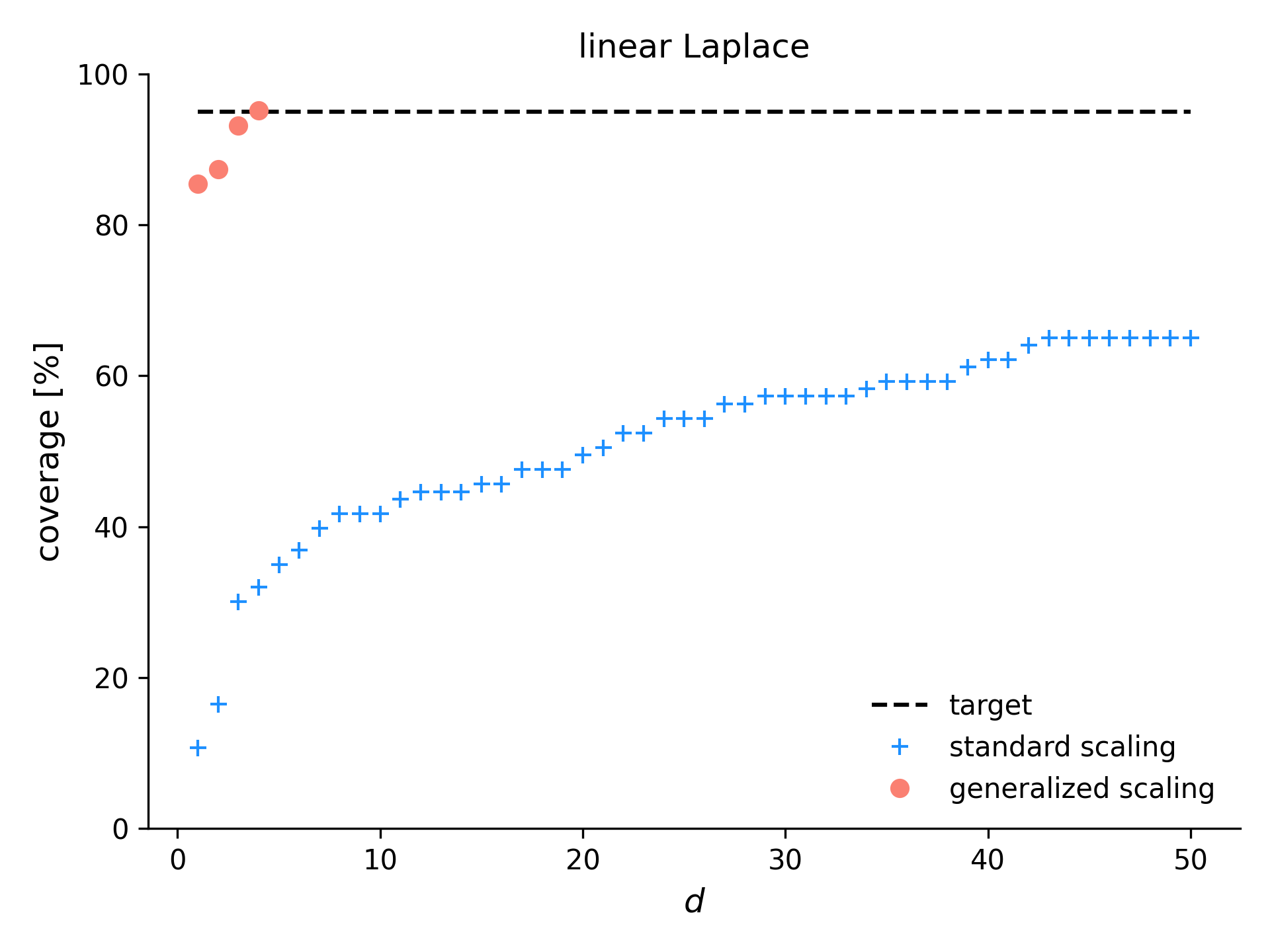}}
\caption{The coverage of the calibration procedure as outlined in Section \ref{sec:d}, versus the active subspace dimension $d$, for both the standard Bayesian scaling $\beta=N\sigma^{-2}$ and the generalized scaling $\beta=\sigma^{-2}$.\label{fig:coverage}}
\end{figure}

{\bf Standard vs generalized scaling.} The coverage for both the generalized and standard Bayesian scaling is plotted versus $d$ in Figure \ref{fig:coverage_a}. In the case of the former, the target coverage was reached at $d=3$, while for the latter it was never reached within the allotted range. In fact, even at $d=50$ only approximately 60\% of the calibration data was covered by the 90\% CIs. 

{\bf Sampled vs linear Laplace.} The same coverage study was conducted with the linear predictive posterior distribution, using again a 95\% confidence interval constructed from $f(x;\theta_0) \pm 2 \sigma_f$. The overall results, see Figure \ref{fig:coverage_b}, are similar compared to the sampled Laplace study. For the generalized scaling, with $d=4$ active dimensions the coverage criterion is met, while the coverage of the standard scaling stagnates.

\begin{figure}[htp]
\centering
\subfigure[Generalized mean-loss scaling, $d=3$. \label{fig:f_concrete_a}]
{\includegraphics[scale=0.38]{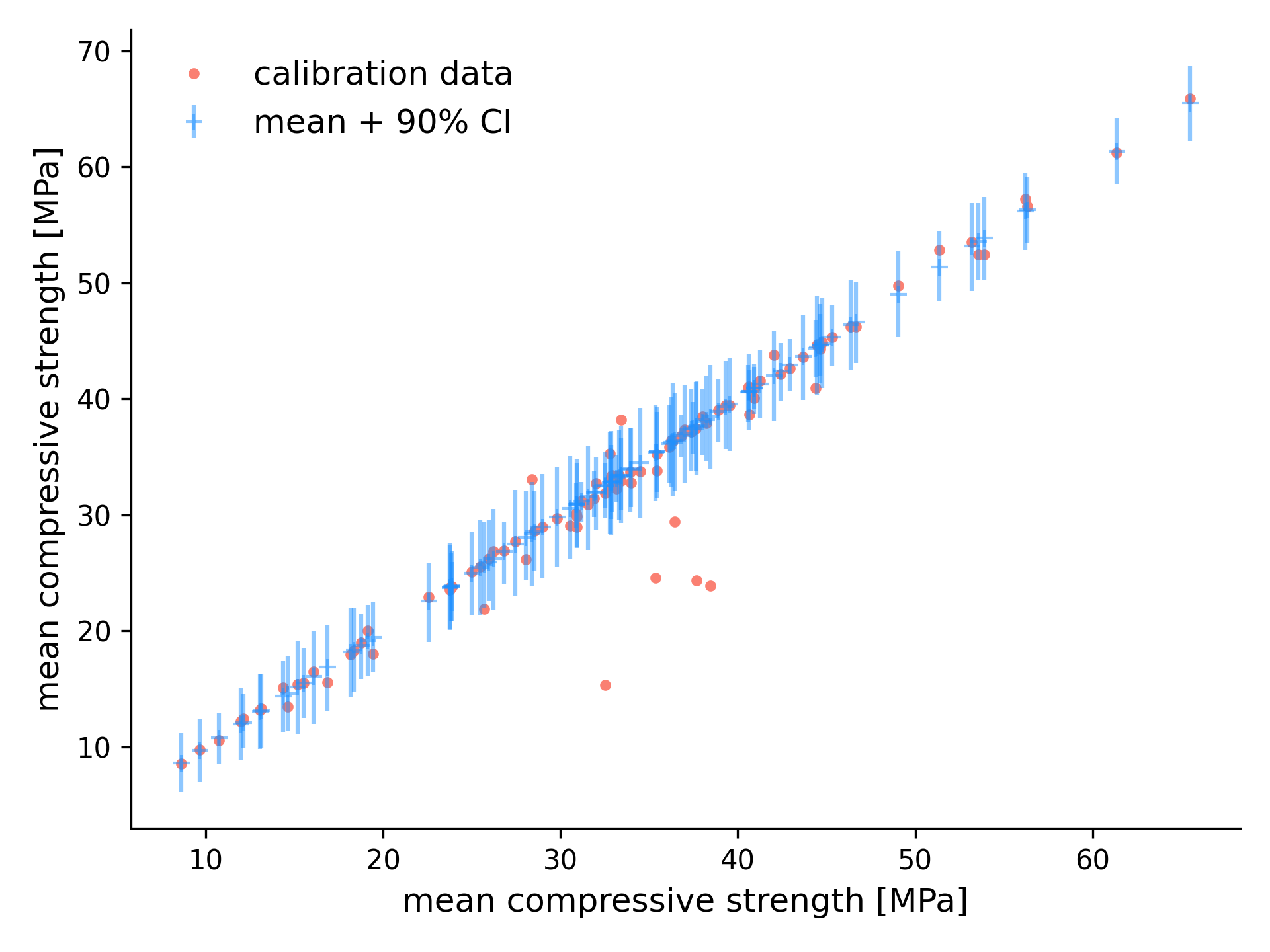}}
\subfigure[Standard Bayes scaling, $d=50$.]
{\includegraphics[scale=0.38]{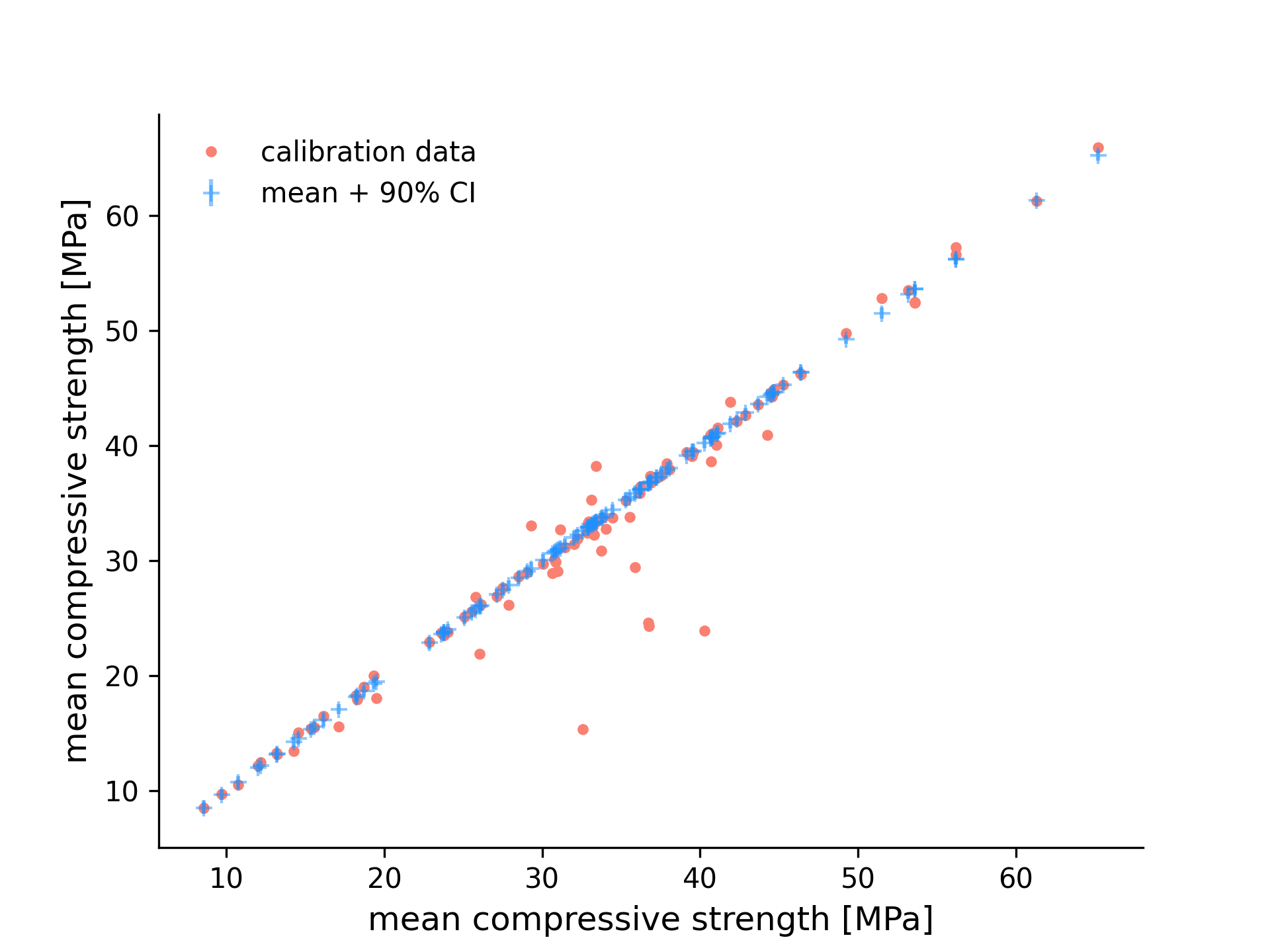}\label{fig:f_concrete_b}}
\caption{The parity plot of the mean compressive strength of concrete, with 90\% CIs. We show the standard and generalized scaling results, both computed with sampled Laplace.\label{fig:f_concrete}}
\end{figure}

Due to the multivariate input $x$, we display the generalized Bayesian posterior predictive results in Figure \ref{fig:f_concrete_a} in the form of a parity plot. Here we show the mean predicted compressive strength versus itself with 90\% sampled Laplace CIs, and the mean predicted compressive strength versus the calibration data. Visually speaking, for the generalized scaling the CIs are consistent, as the mean is generally centered around the validation data, and the 90\% CIs encompass most data without being overly wide. The CIs generated by the standard scaling (Figure \ref{fig:f_concrete_b}) are indeed overconfident. 

\begin{figure}
    \centering
    \includegraphics[width=0.7\linewidth]{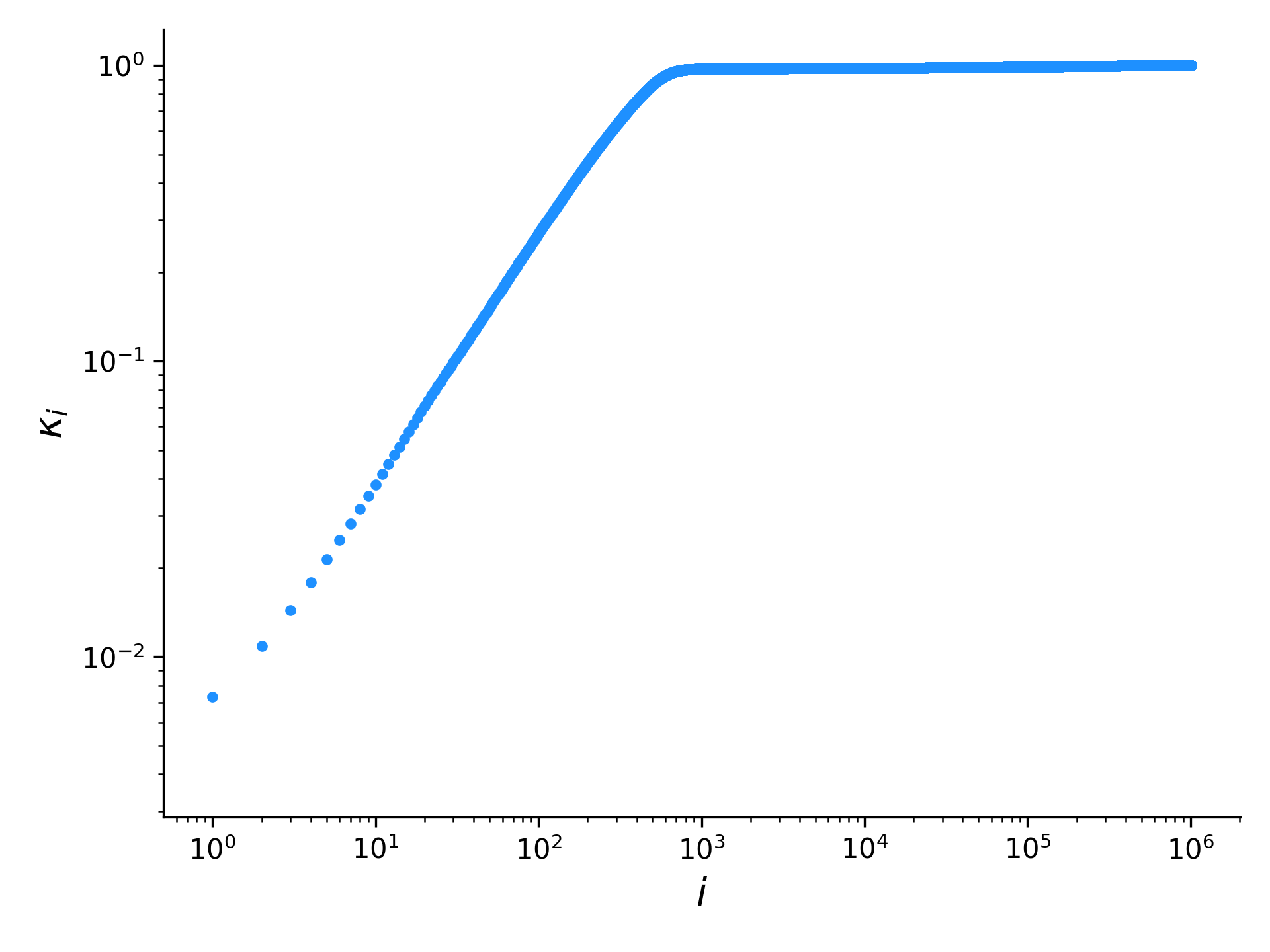}
    \caption{The cumulative (normalized) activity scores for our concrete compression strength feed-forward neural network with approximately $D=10^6$ weights.}
    \label{fig:activity_scores_concrete}
\end{figure}

{\bf Activity scores.} The dominant eigenvectors of $G$ define important directions in parameter space. However, we can also extract information about the importance of individual connections weights. Activity scores, originally derived in \cite{constantine2017global}, are (first-order) sensitivity indices for individual parameters, which can be extracted from the active subspace eigenmodes. Analogously, using the $G$ eigenmodes, we define the activity scores for individual connection weights as:
\begin{align}
    \nu_i = \sum_{j=1}^d \lambda_j p_{j, i}^2, \quad i=1,\cdots,D.
    \label{eq:nu_i}
\end{align}
\noindent
Here $p_{j, i}$ is the $i$-th component of the $j$-th eigenvector of $G$. The activity score $\nu_i$ measures how strongly parameter coordinate $i$ contributes to the retained $d$-dimensional curvature-informed active subspace, weighted by the importance $\lambda_j$ of each active direction. Since the connection weights have no physical interpretation, we will use \eqref{eq:nu_i} primarily to examine if the contribution to the curvature is broadly distributed across the network, or concentrated in a small subset of connection weights. However, rather than plotting individual $\nu_i$, we first sort from largest to smallest and instead plot the cumulative scores
\begin{align}
    \kappa_i = \frac{\sum_{j=1}^i \nu_j}{\sum_{j=1}^D \nu_j},\quad i=1,\cdots,D,
\end{align}
\noindent
in a log-log plot. The results are shown in Figure \ref{fig:activity_scores_concrete}. We see that the 1000 most-sensitive connection weights (just 0.1\% of the total) rapidly saturate and are responsible for most of the curvature sensitivity. In particular, for this case the remaining 99.9\% of weights account for 5\% of the total sensitivity. This indicates that, in addition to being only $d=3$ dimensional, the active subspace is also localised in the high-dimensional weight space.

\subsection{Autoregressive classification for molecular generative modelling \label{sec:reinvent}}

For our final (classification) test case we consider the autoregressive molecular language model from \cite{olivecrona2017molecular}. Here, molecules are represented as simplified molecular-input line-entry specification (SMILES) strings, a sequence of characters corresponding to atoms, generated sequentially by a recurrent neural network trained to predict a discrete distribution for the next token. In particular, we use a 3-layer LSTM network part of the REINVENT tookit (\cite{loeffler2024reinvent}) with 256 neurons per layer and approximately 1.6M weights, trained on SMILES strings from the ChEMBL database (\cite{gaulton2017chembl}). The $M=34$ tokens that make up the SMILES vocabulary and some example ChEMBL molecules can be found in the Appendix. This makes molecular generation a sequence of autoregressive token-level classification problems since, at each step, the network outputs a softmax distribution over the vocabulary, from which the next token is sampled. We study here how the low-rank posterior uncertainty in the network (for both the standard and generalized scaling) affects free-running autoregressive molecular generation. This setting separates the intrinsic stochasticity of the softmax layer from epistemic variability induced by posterior weight perturbations.

\begin{figure}[htp]
    \centering
    \includegraphics[width=0.7\linewidth]{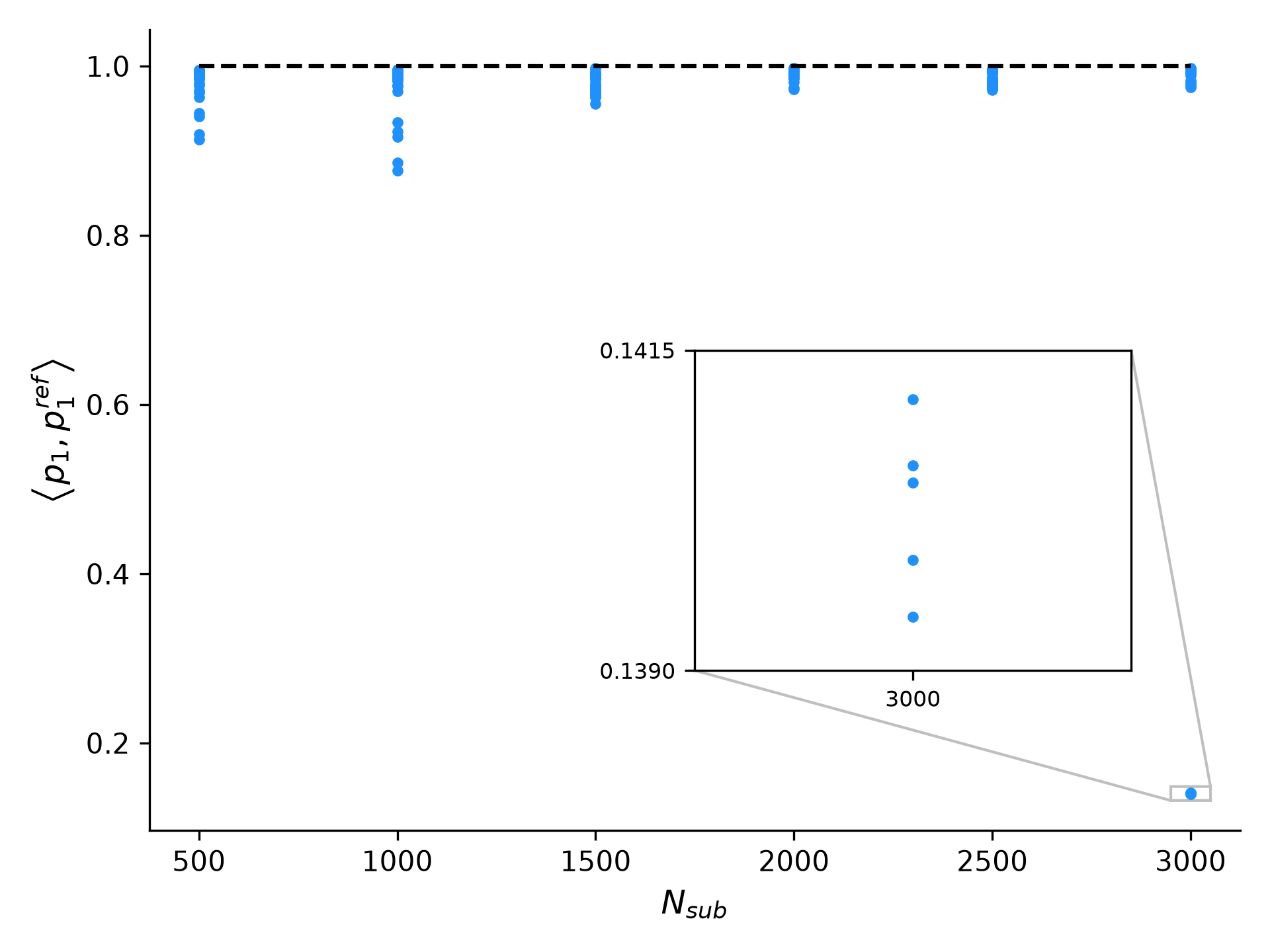}
    \caption{The inner products $\left|\left<p_1, p_1^{ref}\right>\right|$ as a function of $N_{sub}\in\{500,1000,1500,2000,2500,3000\}$, replicated 25 times per $N_{sub}$. The reference vectors $p_1^{ref}$ were computed using $N_{sub}=10.000$ SMILES strings.}
    \label{fig:p1_convergence}
\end{figure}

{\bf Reducing computational cost via subsampling}. The LSTM was trained on $1.25$M SMILES strings, each containing a (varying) number of tokens, which results in a total number of data records of $N=58.7$M. As each $Gv$ product used in the Lanczos algorithm is an empirical average over the dataset (see \eqref{eq:Gv}), we choose to randomly subsample the data for computational efficiency, computing the eigendecomposition of $G$ using $N_{sub}$ SMILES strings instead. Note a similar strategy was employed by \cite{faller2026low}. To examine the effect of this subsampling we plot the (absolute value of) the inner product between the dominant eigenvector $p_{1}$ and a reference vector $p_1^{ref}$ as a function of $N_{sub}$ in Figure \ref{fig:p1_convergence}. This reference vector was computed using $N_{sub} = 10k$, and we replicate this procedure $N_{rep}=5$ times at each subsampling to account for sampling noise, creating 25 possible inner products per $N_{sub}$. While clear variation is present in the replica inner products for $N_{sub} = 500$ and $N_{sub} = 1000$, $\left|\left<p_1, p_1^{ref}\right>\right|$ soon clusters near 1, indicating that reasonable alignment with the reference eigenvector can be obtained with a small fraction of the total dataset. That said, we note that there is an obvious outlier for $N_{sub}=3000$. This is most-likely caused by a $p_1$ which failed to converge during the Lanczos algorithm, such that all 5 corresponding $\left|\left<p_1, p_1^{ref}\right>\right|$ cluster near 0.14 in this case. One simple way to flag such outliers for a given $N_{sub}$ is computing the $N_{rep}\times N_{rep}$ matrix with entries $\left|\left<p_{1,i}, p_{1,j}\right>\right|$, $i,j=1,\cdots,N_{rep}$, $p_{1,i}$ being the i-th replica eigenvector. Ill-converged vectors will cause entries away from 1. All subsequent analysis was performed using $N_{sub}=2000$ SMILES, corresponding to $N=93k$ data points.

{\bf Standard vs generalized scaling.} Our first results concern the output validity, as not all generated molecules are guaranteed to be chemically correct compounds. REINVENT verifies the validity in a post processing step using the RDKit \cite{landrum2026rdkit}. Note that validity here means RDKit can parse the SMILES string into a molecular graph and the graph passes basic chemical sanitization checks. It does not imply drug-likeness, synthetic accessibility, or chemical plausibility. Table \ref{tab:validity} shows the number of valid molecules out of a set of 10k generated SMILES strings for the baseline REINVENT model (i.e.\ with softmax-only uncertainty and weights fixed at $\theta_0$), and the perturbed model Laplace models with standard and generalized scaling. When varying the active-subspace dimension from 1 to 3, the standard scaling does not result in a notable change in validity compared to the baseline model. For the generalized scaling, validity notably drops.

\begin{table}[ht]
\centering
\caption{Number of valid SMILES strings (out of 10k). \label{tab:validity}}
\begin{tabular}{cccc}
\hline\hline
 &  baseline & Laplace (std.\ scaling) & Laplace (gen.\ scaling)\\
\hline
$d=1$ & 8692 & 8709 & 6863 \\
$d=2$ & 8692  & 8753  & 5279 \\
$d=3$ & 8692  & 8732 & 4435 \\
\hline\hline
\end{tabular}
\end{table}

\begin{figure}
    \centering
    \includegraphics[width=0.8\linewidth]{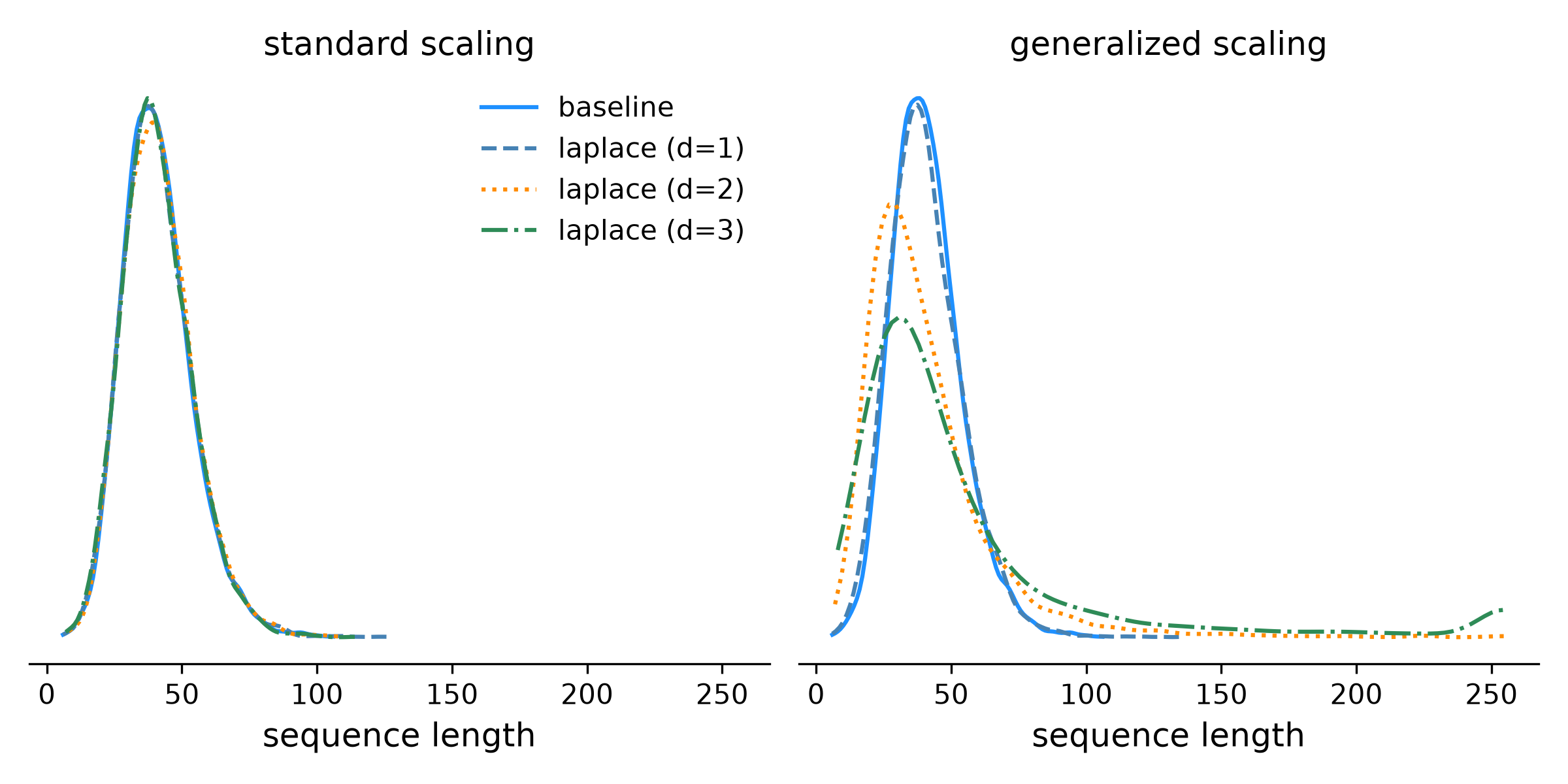}
    \caption{Kernel density estimates of the sequence length for the standard and generalized scaling with $d\in\{1,2,3\}$. The maximum allowed sequence length is 256.}
    \label{fig:seq_length}
\end{figure}

\begin{figure}
    \centering
    \subfigure[Molecular weight distributions.]
    {\includegraphics[width=0.8\linewidth]{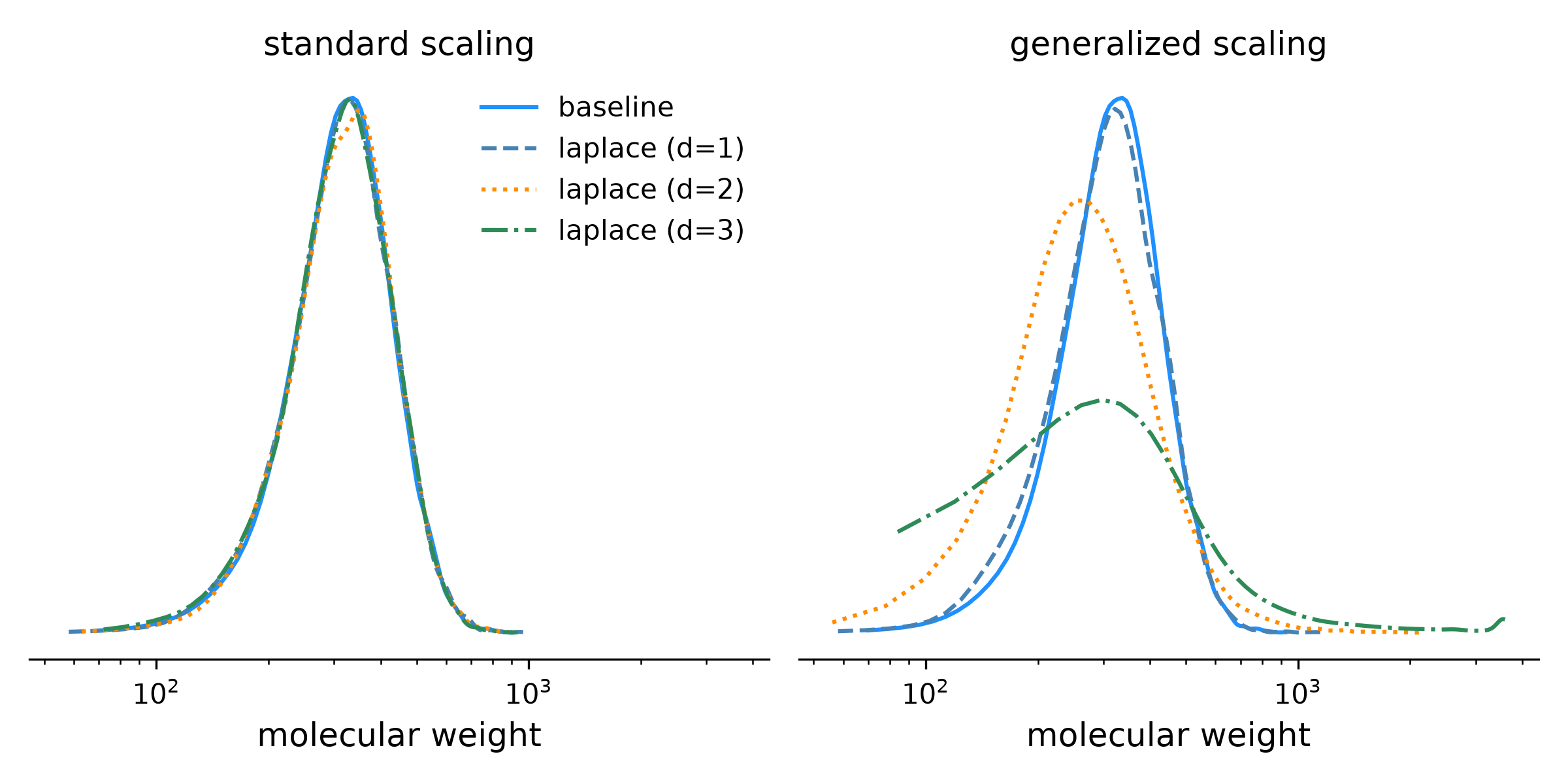}}
    \subfigure[SA score distributions.]
    {\includegraphics[width=0.8\linewidth]{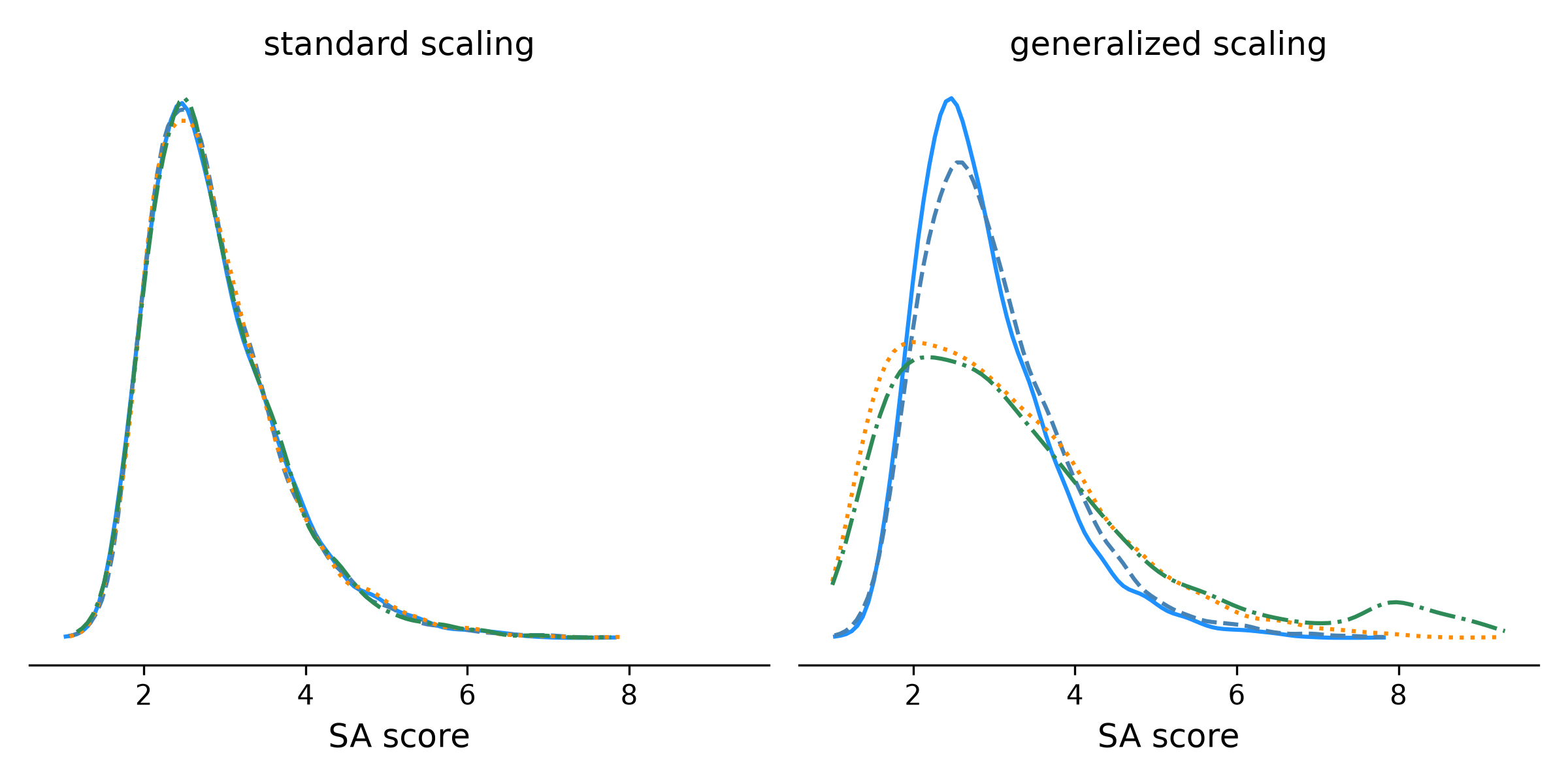}}
    \subfigure[SlogP distributions.]
    {\includegraphics[width=0.8\linewidth]{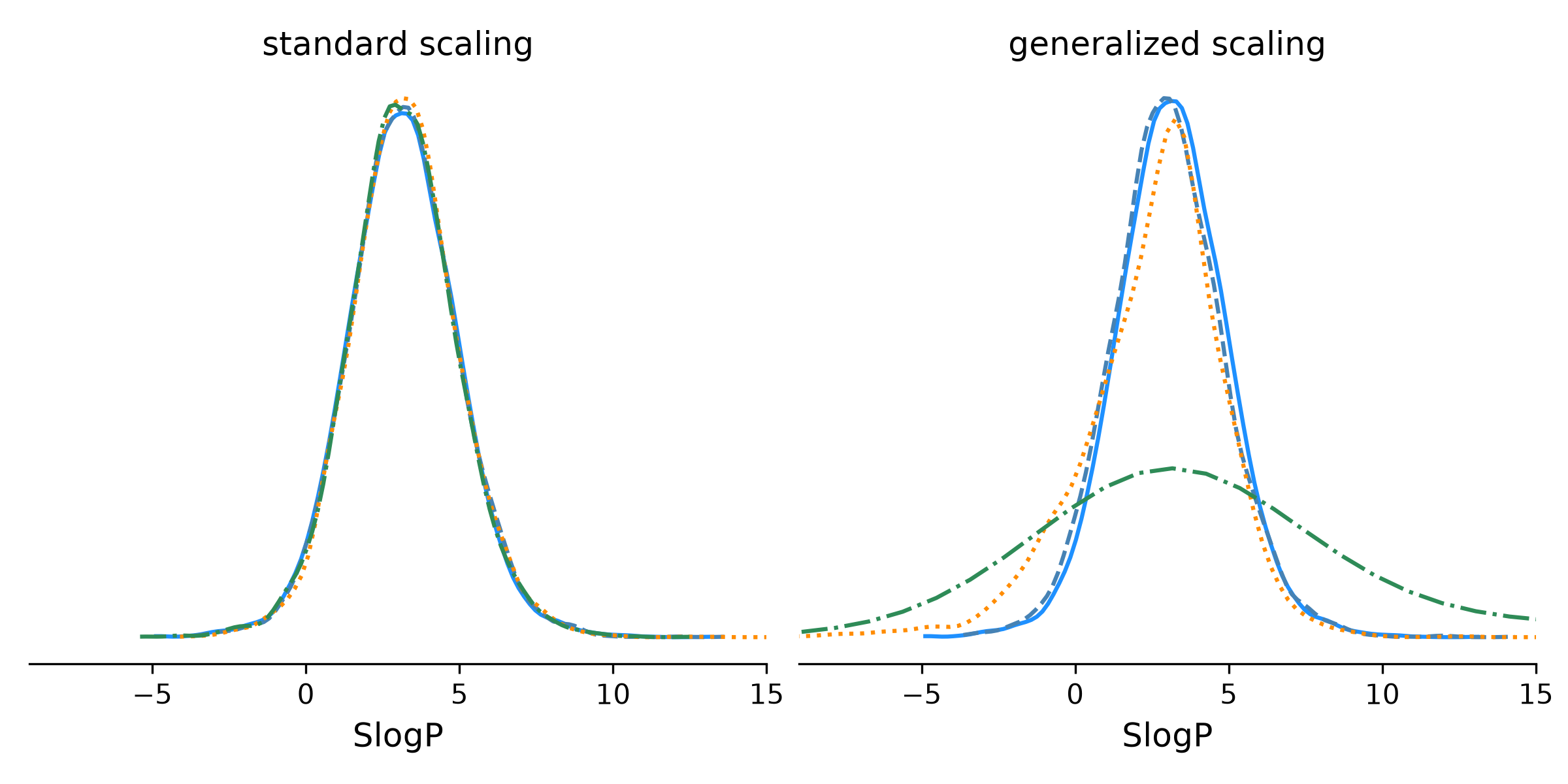}}
    \caption{Kernel density estimates of the QoIs for the standard and generalized scaling with $d\in\{1,2,3\}$. \label{fig:kdes}}
\end{figure}

The following results were obtained using only valid SMILES strings. Figure \ref{fig:seq_length} shows the kernel density estimates of the sequence length of the molecules (i.e.\ the number of tokens), computed from 10k generated SMILES strings. Once again only the generalized scaling shows a departure from the baseline results. Especially for $d=3$ the generator produces longer SMILES strings, with an increased probability that the maximum allowed sequence length of 256 is reached. Finally, we examine the effect of the Laplace active subspaces on the distribution of several Quantities of Interest (QoIs), described in Table \ref{tab:qoi}. The corresponding distribution are found in Figure \ref{fig:kdes}. 

\begin{table}[ht]
\centering
\caption{Molecular quantities of interest used to compare generated SMILES distributions within REINVENT.}
\label{tab:qoi}
\small
\begin{tabular}{ll}
\hline
Quantity & Description  \\
\hline\hline
Molecular weight  &
Molecular size based on atomic masses, large values indicate large molecules. \\
SA score & Synthetic Accessibility score, higher values indicate more difficult synthesis. \\
SlogP & Estimate of how fat-soluble a molecule is. \\
\hline\hline
\end{tabular}
\end{table}

\begin{figure}
    \centering
    \includegraphics[width=0.7\linewidth]{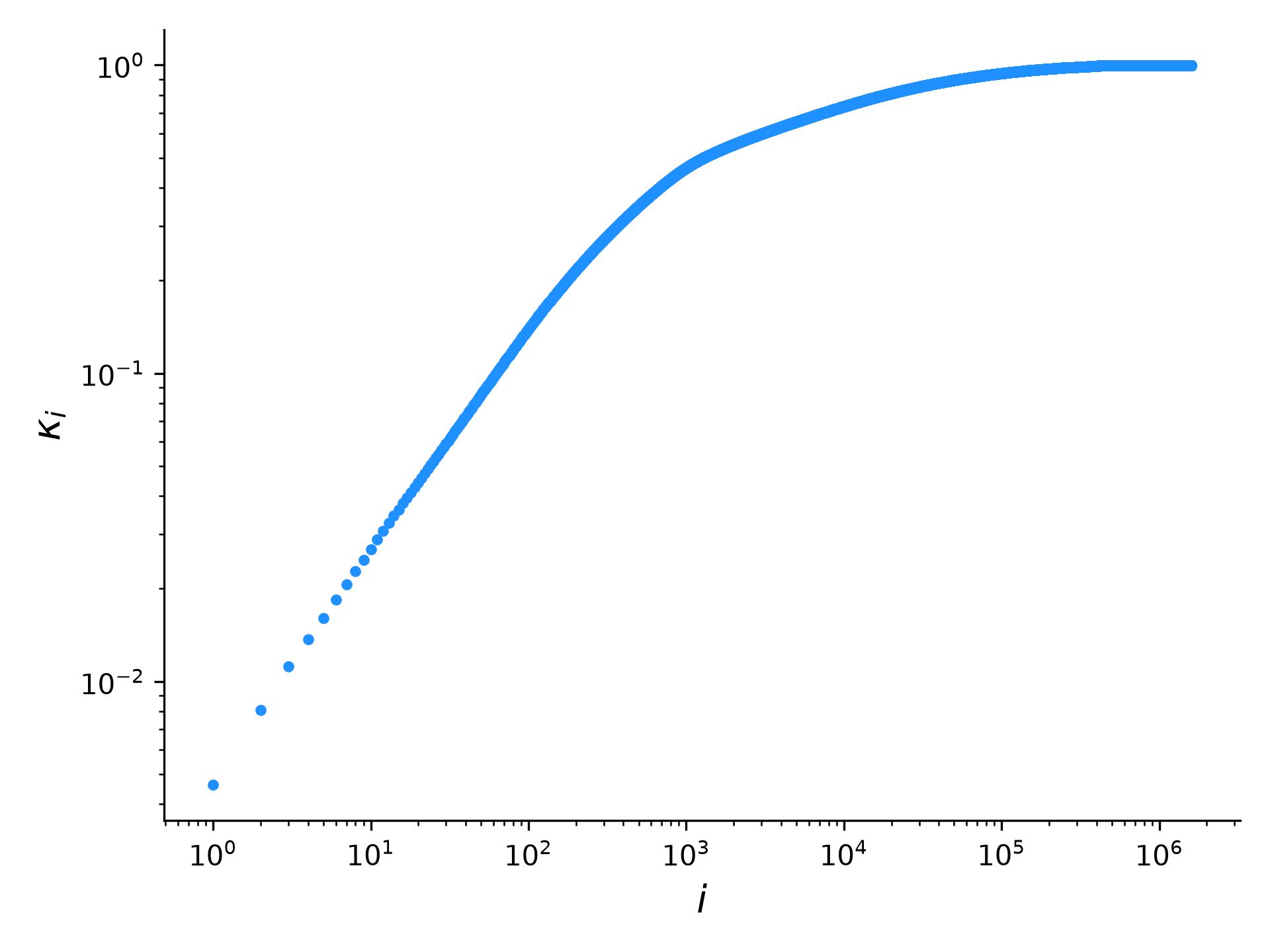}
    \caption{The cumulative (normalized) activity scores for the REINVENT LSTM with approximately 1.6M weights. Computed using $d=2$. \label{fig:activity_scores_reinvent}}
\end{figure}

{\bf Activity scores}. The cumulative activity scores $\kappa_i$ are shown in Figure \ref{fig:activity_scores_reinvent}, using $d=2$. Note the curve saturates less quickly compared to the concrete compression strength case. Still, the first 7.6\% of the activity scores already account for 95\% of the total sensitivity.

\section{Discussion}

We now discuss our findings and contrast our proposed method with related work.

\subsection{Posterior scaling and active-subspace dimension}

The results show that the scaling of the generalized posterior plays a central role in the quality and efficiency of our low-rank approximation. Under the standard Bayesian scaling, the leading active directions become increasingly contracted with the number of training data, as $\sigma^2_j=(\beta\lambda_j + \sigma^{-2}_0)^{-1}$ with $\beta=N\sigma^{-2}$ or $\beta=N$ for regression or classification respectively. This forces coverage calibration to retain many weak-curvature directions. 

Note that this does not mean the standard scaling is intrinsically defective, since posterior contraction with increasing data is an expected feature of the standard Bayesian update. The standard Bayesian scaling and generalized mean-loss scaling define do different posterior objects, and in our specific context of low-rank posterior UQ for overparameterized neural networks, the generalized scaling can provide a more effective posterior approximation.

\begin{figure}
\centering
\subfigure[$\beta = \sigma^{-2}$, $N=50$ \label{fig:contraction_a}]
{\includegraphics[scale=0.38]{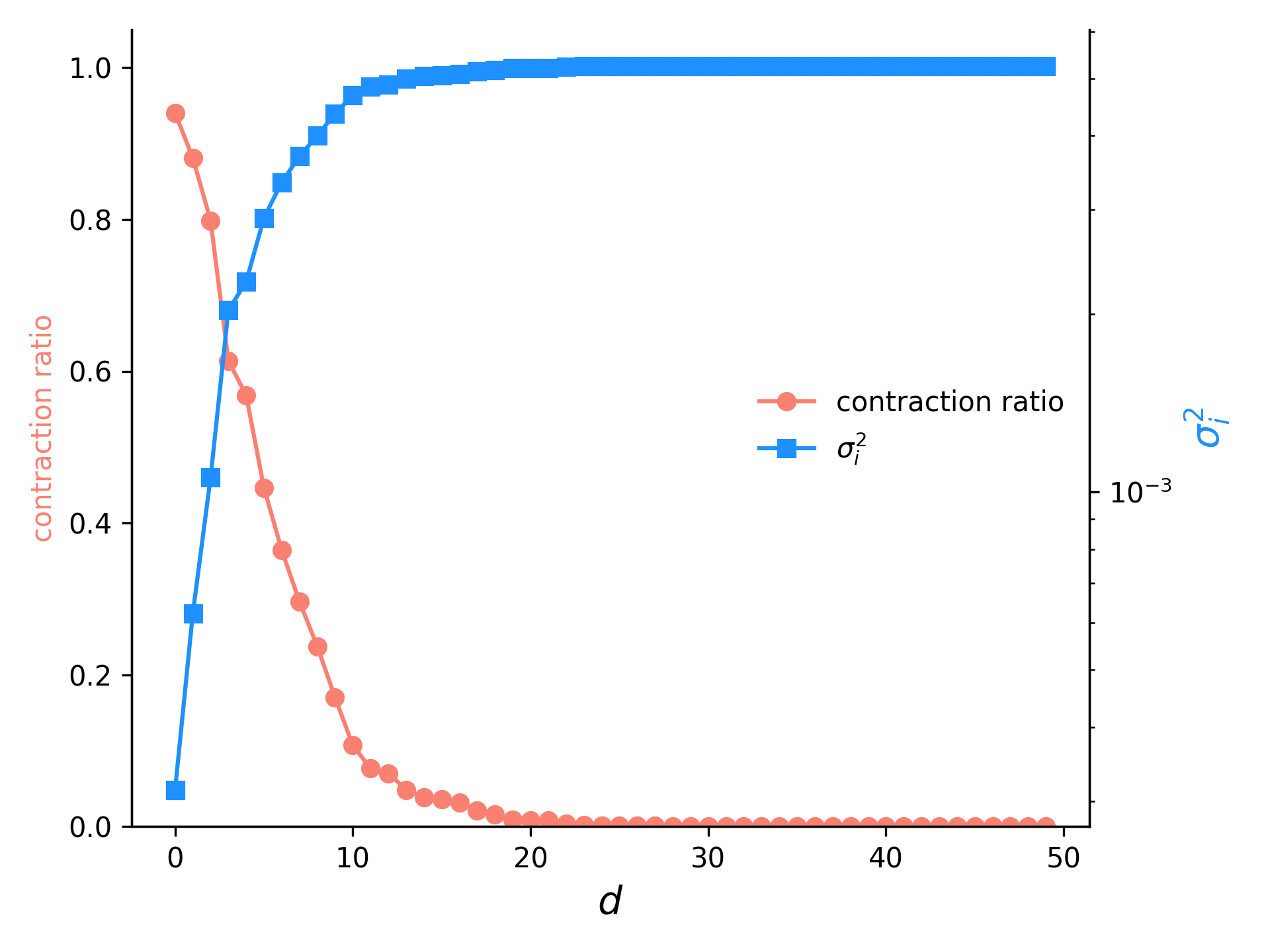}}
\subfigure[$\beta = N\sigma^{-2}$, $N=50$ \label{fig:contraction_b}]
{\includegraphics[scale=0.38]{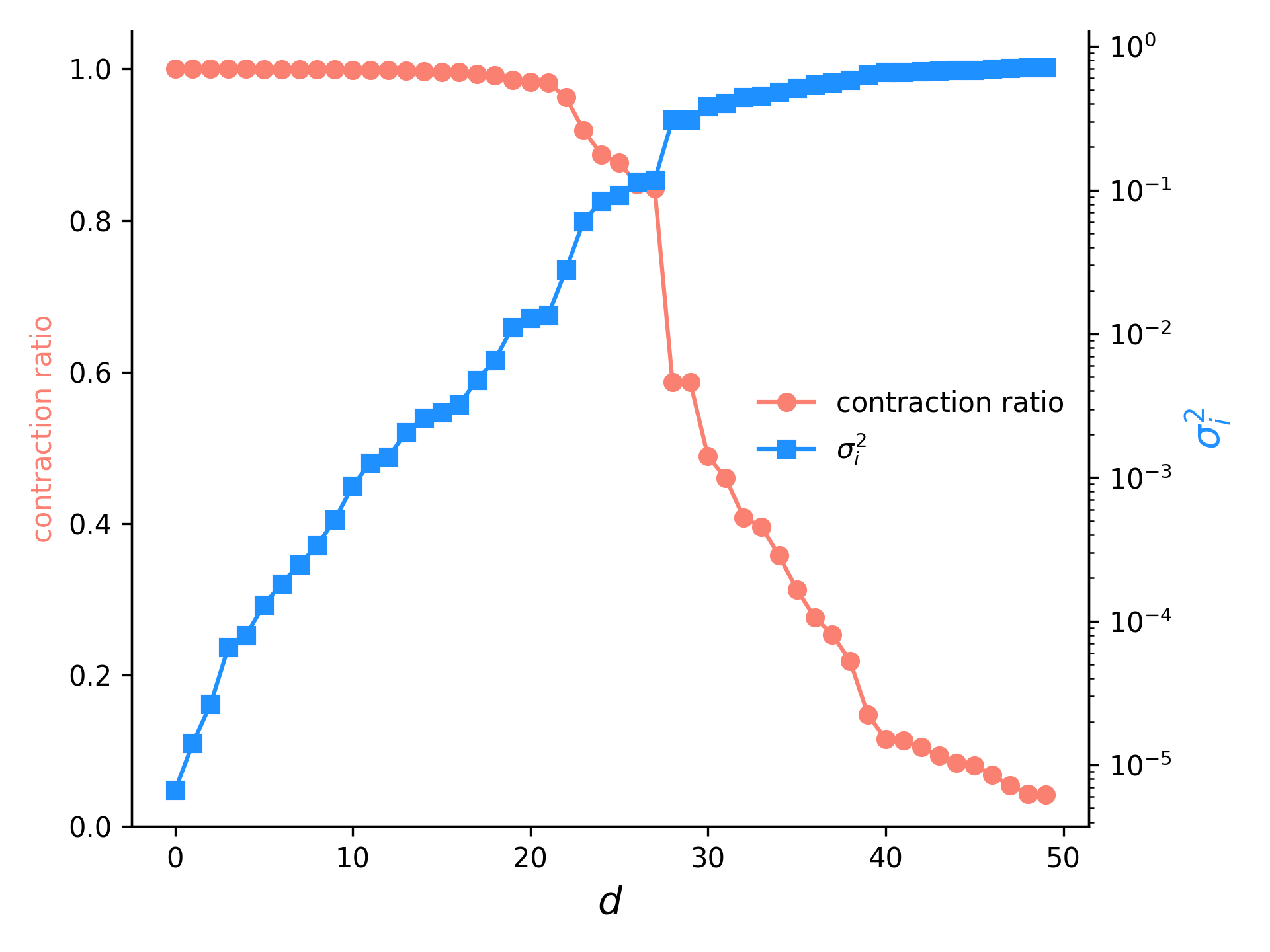}}
\subfigure[$\beta = \sigma^{-2}$, $N=1000$ \label{fig:contraction_c}]
{\includegraphics[scale=0.38]{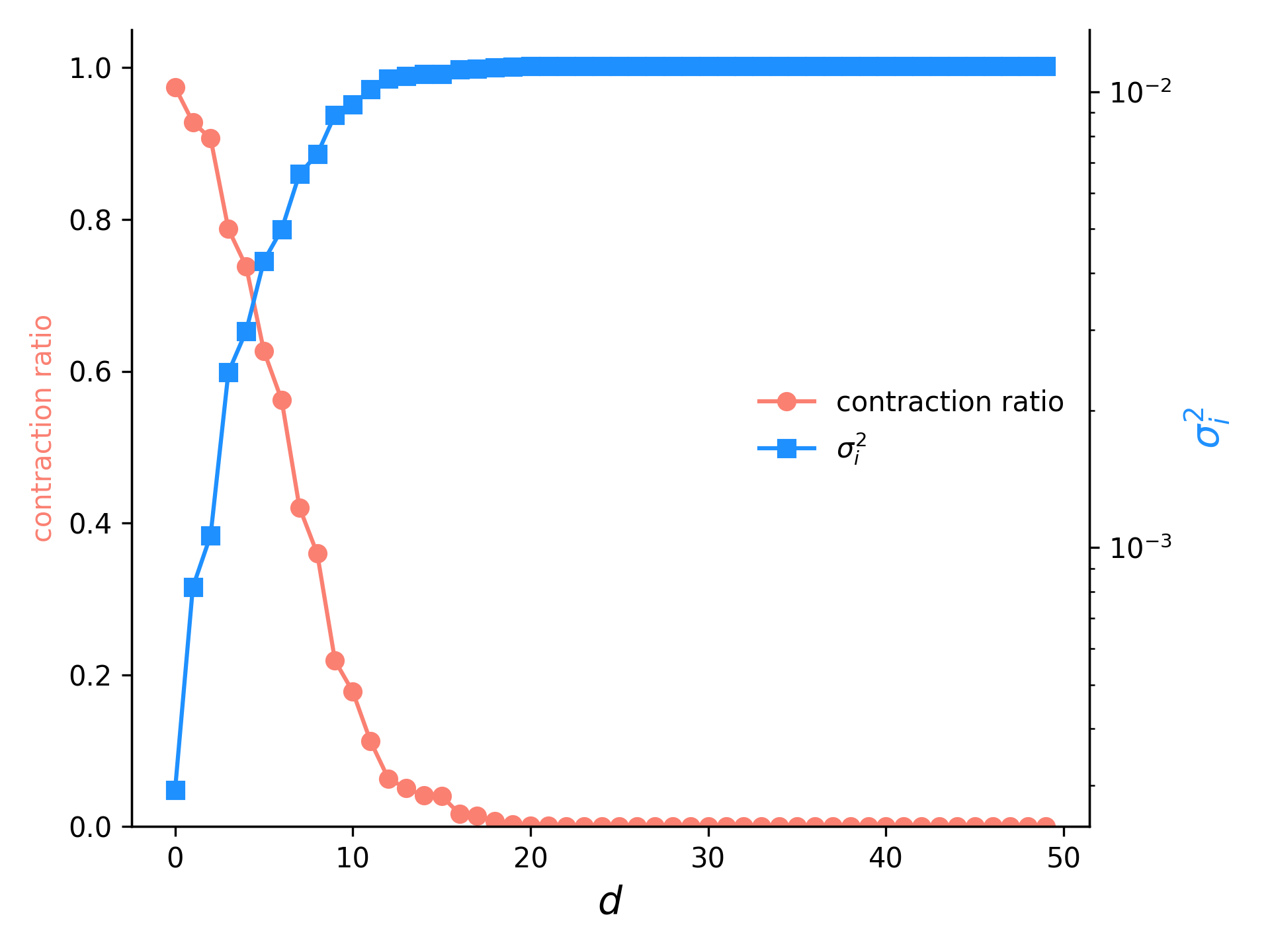}}
\subfigure[$\beta = N\sigma^{-2}$, $N=1000$ \label{fig:contraction_d}]
{\includegraphics[scale=0.38]{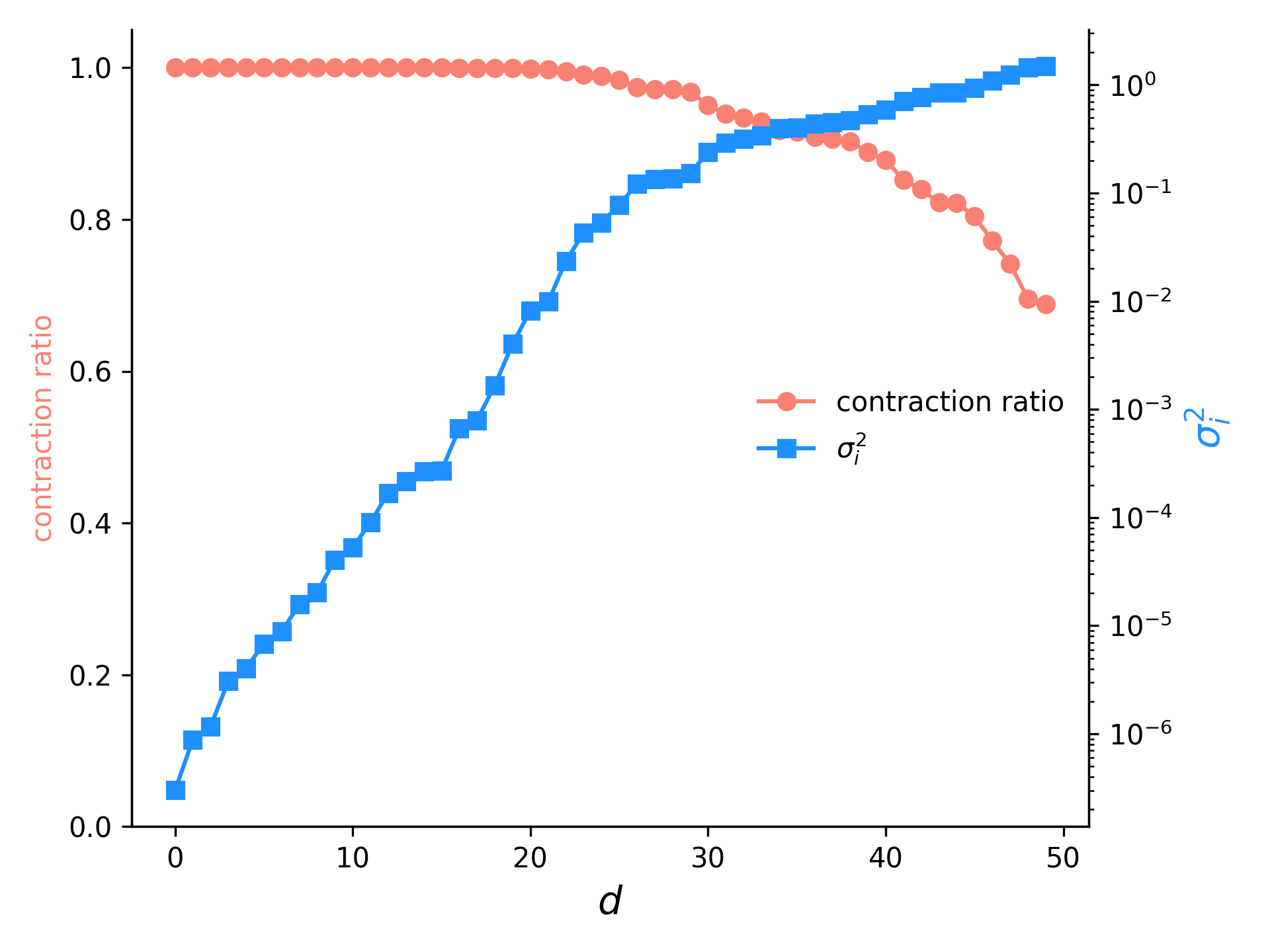}}
\caption{The contraction ratio \eqref{eq:c_i} and posterior variances \eqref{eq:post_var} for the generalized (left column) and standard (right column) Bayesian scaling, for both $N=50$ and $N=1000$. These results were computed using the analytic regression function \eqref{eq:f_analytic}. \label{fig:contraction}}
\end{figure}

To further discuss this behaviour, consider Figure \ref{fig:contraction}. For the analytic regression function \eqref{eq:f_analytic}, we plot the posterior variances and the contraction ratio;
\begin{align}
c_i := \frac{\sigma^2_0 - \sigma^2_i}{\sigma^2_0} = \frac{\beta \lambda_i}{ \beta \lambda_i + \frac{1}{\sigma^2_0}},
\label{eq:c_i}
\end{align}
for which $c_i \approx 1$ implies that the $i$-th direction is strongly data-informed (since then 
$\sigma^2_i\rightarrow 0$), and values close to zero denote prior-dominated directions ($\sigma^2_i\rightarrow\sigma^2_0$). Under the standard Bayesian scaling (Figures \ref{fig:contraction_b}-\ref{fig:contraction_d}), while the leading directions are almost fully contracted (posterior dominated), the posterior variances remain close to zero over much of the leading spectrum, due to the scaling with $N$: $\sigma^{2}_i\approx 1/(N\sigma^{-2}\lambda_i)$ when $c_i \approx 1$. Coverage calibration therefore selects directions much deeper in the spectrum, where the posterior variances increase rapidly, as the $\sigma^2_i$ span multiple orders of magnitude. Consequently, nominal coverage is obtained only after adding directions that introduce substantial parameter-space variability along weak-curvature modes. This effect becomes more pronounced with more training data; in Figure \ref{fig:contraction_d} the variance along the leading active directions is reduced by an order of magnitude compared to Figure \ref{fig:contraction_b}. As a result, $d=43$ was required to meet the coverage criterion, which exacerbates the aforementioned problems of incorporating weak curvature modes with high variance. In the non-linear network, these larger perturbations may move samples outside the local region in which the Laplace approximation is accurate, resulting in wider and less stable predictive intervals, as shown in Figure \ref{fig:N1000}. A better mean posterior prediction can be obtained with a lower $d$, at the cost of overconfident CIs, or by utilizing the linear predictive distribution. Such overconfident CIs were also observed with the UCI concrete compressive strength network with standard scaling, where 50 curvature directions were insufficient to achieve coverage. The slow increase displayed in Figure \ref{fig:coverage} also suggests a significantly higher subspace dimension is required to achieve target coverage. Moreover, the retained eigenvalue spectrum spans several orders of magnitude, indicating that the additional directions correspond to genuinely low-curvature modes.

While the linear predictive distribution does remove the non-linear sampling artifacts when they occur, the excessive posterior contraction with $N$ induced by the standard Bayesian scaling is still present for linear Laplace. This is clear from Figures \ref{fig:f_lin}-\ref{fig:f_concrete}. In contrast, under the generalized Bayesian scaling, sufficient uncertainty is retained in the leading active directions, regardless of $N$ or the posterior predictive method, and desired coverage was obtained with only $d=4$ or less in all cases. Thus, the uncertainty is concentrated in a small number of coherent active directions rather than spread across many weak-curvature modes.

The increase in the selected active subspace dimension under the standard Bayesian scaling is also computationally consequential. Although $N=1000$ is not large by modern machine-learning standards, the standard scaling already requires a substantially higher posterior rank to achieve the desired coverage compared to $N=50$, which is a problem that will escalate with larger $N$. That said, as we showed in Section \ref{sec:reinvent}, for large datasets one could subsample the data when computing the eigenmodes of $G$. Still, computing these modes under the standard scaling will require a larger $d$, and therefore more Lanczos iterations (see Section \ref{sec:matrix_free}). Thus, the standard Bayesian scaling not only shifts uncertainty into weaker-curvature directions, but also increases the computational cost of the low-rank posterior approximation compared to the generalized scaling. In contrast, the generalized Bayesian mean-loss scaling achieves calibrated and sharp confidence intervals with a much smaller active subspace. The proposed approach therefore provides a practical route to calibrated predictive uncertainty for regression-based neural networks.

Similarly, for our autoregressive classification case, the standard scaling resulted in distributions that do not deviate from the baseline REINVENT model in any notable fashion, see Figure \ref{fig:kdes}. The low-rank Laplace perturbations with $d=1,2$ appear to add meaningful variation to the generated molecular distribution while remaining broadly within the regime learned by the pretrained generator. In contrast, $d=3$ clearly crosses a stability boundary. The generator can start to produce some long (carbon-rich), formally valid yet chemically implausible molecules with extreme SlogP values. Lipinski’s Rule of Five \cite{lipinski1997experimental} is a rule of thumb used in drug discovery to evaluate if a chemical compound has the right properties to make it an orally active drug in humans. One violating condition of this rule is an SlogP value greater than 5, which occurs at a significantly higher probability once $d=3$. This suggests that the addition of third active direction pushes the autoregressive dynamics outside the chemically meaningful region of the model.

Overall, under standard Bayesian scaling, the low-rank Laplace generator remains close to the pretrained baseline for all tested subspace dimensions. Valid-molecule QoI distributions and sequence lengths are not meaningfully affected, indicating that the posterior samples remain in a local generative regime. The reason behind this is the same as in the regression case, and again will be become even more pronounced with increasing (subsampled) data size.  In contrast, under generalized scaling, increasing the active-subspace dimension produces progressively stronger deviations from the baseline. The most pronounced effect is in the sequence-length distribution, where a higher $d$ introduces a long right tail, suggesting that generalized posterior perturbations affect the learned stopping behaviour. The same scaling also broadens the QoI distributions, indicating increased chemical exploration at the cost of generative stability. 

However, this was only a sensitivity study of the effect of posterior weight perturbations on free running molecule generation. Our pretrained REINVENT model is already reasonably calibrated at the token level, and we therefore did not expect the additional uncertainty of the Laplace posterior to improve the predictive distribution. That said, low-rank Laplace methods have also been proposed for fine-tuned large language models (LLMs). For instance, the low-rank adaptation (LoRA) method of \cite{hu2021lora} is a means to fine tune an LLM by introducing low-rank updates to selected weight matrices. This deterministic fine-tuning can make the model poorly calibrated, often overconfident. Laplace-LoRA, introduced by \cite{yang2024bayesian} places a Laplace posterior over the fine-tuned LoRA parameters to induce variability in the logits/predicted probabilities. The prior precision is tuned using gradient descent, maximizing the Laplace evidence, such that the posterior predictive distribution becomes better calibrated and achieves lower negative log likelihood. Extending our generalized-Bayes active-subspace Laplace approximation to fine-tuned language models is therefore a promising direction for future work. This would shift the role of the Laplace approximation from probing generative sensitivity, as in the present REINVENT experiments, toward correcting overconfidence in fine-tuned autoregressive models.

Finally, the activity scores from Figures \ref{fig:activity_scores_concrete} and \ref{fig:activity_scores_reinvent} show that the active subspace is concentrated in a very small subset of all connection weights, relative to the total weight space. However, the active subspace $w=P^T\theta$ itself provides an even more compact representation, as it is a linear combination of connection weights instead of a collection of individual weights. Similar observations have been made in physics-based models, see e.g.\ \cite{edeling2024global}. 


\subsection{Related work}

In recent years several contributions to the problem of uncertainty quantification for neural networks have been made, see \cite{psaros2023uncertainty} or \cite{wang2025review} for survey articles. Here we discuss those most relevant to our work.

Our work is related to the subspace inference framework of \cite{izmailov2020subspace}, where Bayesian model averaging is performed within low-dimensional parameter subspaces constructed from random projections, SGD-trajectory PCA, or mode-connecting curves. In contrast, we construct the subspace from the dominant generalized Gauss--Newton curvature directions of the empirical mean loss. In \cite{izmailov2020subspace} posterior tempering is also used as a calibration device, although here the temperature is a hyper parameter selected by cross-validation, rather than being fixed by the loss function used in generalized Bayes. Our curvature-based setting exposes the different scaling issue of increasingly contracted leading active directions with $N$ for the standard scaling. The generalized Bayesian mean-loss scaling avoids this purely data-size-driven contraction. A further distinction is that our generalized Laplace approximation yields a closed-form Gaussian posterior in the active subspace, avoiding Markov-Chain Monte Carlo sampling or variational optimization within the subspace. 

Most closely related is the low-rank Laplace subspace construction of \cite{faller2026low}, whose authors derive an optimal low-dimensional subspace model for approximating the predictive output covariance of a full Laplace approximation and propose a scalable compute strategy. Their construction is formulated in the linearized Laplace setting discussed previously. This linearization yields a closed-form Gaussian on both the networks weights and the output, which \cite{faller2026low} use to find the subspace projection matrix $P$ by minimizing the discrepancy between the output covariance of the subspace model and that of the full Laplace approximation. However, the exact construction requires access to quantities that are generally inaccessible for large neural networks, including the full (GGN) curvature matrix and an eigendecomposition of the full output covariance. In practice, scalable variants therefore rely on additional approximations such as diagonal or Kronecker-factored curvature estimates \cite{grosse2016kronecker}.

Our goal is different. Rather than computing the output covariance of the full Laplace approximation, we construct a generalized-Bayes Laplace approximation directly in a curvature informed active subspace. The posterior samples are propagated through the non-linear network itself rather than through a linearized predictive model. This distinction is important since linearized Laplace methods keep the posterior predictive mean anchored at the MAP prediction (see section \ref{sec:pred}), and can thereby avoid the underfitting behaviour observed when non-linear weight samples are propagated through the network. In our non-linear predictive setting, the size and coherence of the sampled parameter perturbations are therefore central; if the calibrated posterior requires many weak-curvature directions, the resulting samples may leave the local region in which the Laplace approximation is reliable. The generalized Bayesian scaling mitigates this effect by achieving calibrated coverage with a much smaller active subspace, thereby reducing both non-linear sampling artifacts and the number of curvature-vector products required.

Also, the authors of \cite{faller2026low} shows that a low-dimensional subspace can approximate the full predictive covariance of the linearized Laplace approximation, but the required subspace dimensions are still in the tens to hundreds in their experiments. While this is tiny compared to the size of the full weight space, our generalized-Bayes active subspace often achieves calibrated coverage with only a handful of curvature directions, preserving the intended low-rank character of the approximation. 

There are a number of other contributions that use the Laplace approximation of the posterior. The authors of \cite{ritter2018} proposed a scalable Laplace approximation for neural networks using the aforementioned Kronecker-factored curvature estimates of \cite{grosse2016kronecker}. This approach improves scalability by imposing a layer-wise Kronecker structure on the curvature, whereas our method obtains scalability by restricting the posterior to a low-rank generalized Gauss-Newton active subspace. Consequently, our approach does not require layer-wise Kronecker independence assumptions. The broader practical relevance of Laplace approximations for Bayesian deep learning has been demonstrated by \cite{daxberger2021laplace}, whose authors review and benchmark several scalable Laplace variants and provide a dedicated laplace PyTorch library.

Related active-subspace approaches based on variational inference include \cite{jantre2024learning} and \cite{Nafiz_Abeer_2024}. These methods draw on the classical active-subspace construction, in which the subspace is identified from an uncentered covariance of model gradients \cite{constantine2014active}. This matrix is typically averaged over a prescribed prior weight distribution. In contrast, our active subspace can be viewed as a Monte Carlo approximation of a data-averaged local curvature matrix. Specifically, the generalized Gauss--Newton matrix is an empirical average over the training data, evaluated at the pretrained weights. Thus, unlike prior-averaged active-subspace constructions, our subspace is tied directly to the local data-fit geometry that also defines the generalized Laplace approximation.




\section{Methods\label{sec:glas}}

We now describe the methodological details of the proposed low-rank generalized Laplace framework for uncertainty quantification in neural networks. 

\subsection{Statistics}

Our method starts from a generalized Bayesian posterior defined through the empirical mean loss, rather than the summed negative log-likelihood used in standard Bayesian analysis. Around a pretrained set of weights, we then construct a Laplace approximation whose covariance is determined by the local curvature of the loss. Since the full Hessian is intractable and may be indefinite, we replace it by a positive semi-definite generalized Gauss-Newton approximation \cite{kunstner2019limitations} and compute its dominant eigenspace using matrix-free products computed with automatic differentiation. The resulting Gaussian posterior is restricted to this low-dimensional active subspace, with the prior variance calibrated by an empirical generalized-Bayes evidence criterion. We end with a discussion on posterior sampling strategies.

\subsubsection{Generalized Bayes formulation \label{sec:gen_bayes}}

To define a posterior update, a mechanism to connect the parameters to the data is required. In a generalized Bayesian setting this is achieved via a loss function rather than through a traditional likelihood function. This allows us to construct a generalized posterior based on the empirical mean loss while retaining the familiar structure of a prior-to-posterior belief update. We define the mean loss function as
\begin{align}
\mathcal{L}_N(\theta, y) = \frac{1}{N}\sum_{n=1}^N l_n(\theta, y_n).
\label{eq:loss}
\end{align}
\noindent
We will consider:
\begin{align}
l_n = \frac{1}{2}\sum_{m=1}^M \left(y_n^{(m)} - f^{(m)}_n\right)^2 \quad\mathrm{and}\quad l_n = -\sum_{m=1}^M y^{(m)}_m\log p_n^{(m)},
\end{align}
\noindent 
i.e.\ squared-loss and cross-entropy loss terms for regression and classification respectively. In the former $y_n\in\mathbb{R}^M$ are the (noisy) observations of the true process, whereas in the latter $y_n$ is the $M$-dimensional one-hot encoded vector with $y_n^{(m)}=1$ for $m=c$ and $y_n^{(m)}=0$ for $m\neq c$, $c$ being the index of the correct class. Here, the neural-network outputs $f_n^{(m)}:=f^{(m)}(x_n;\theta)$ are the logits that determine the softmax probabilities
\begin{align}
    p_n^{(m)} = \frac{\exp(f_n^{(m)})}{\sum_{j}\exp(f_n^{(j)})}.
    \label{eq:softmax}
\end{align}

The generalized Bayesian (unnormalized) posterior is given by \cite{bissiri2016general} as:
\begin{align}
\tilde\pi_\beta(\theta\mid y) = \exp\left[-\beta\mathcal{L}_N(\theta, y)\right]p(\theta),
\label{eq:gen_post}
\end{align}
where we assume the following prior:
\begin{align}
p(\theta)=\mathcal{N}\left(0, \sigma^2_0 I_D\right).
\end{align}
\noindent
Here, $I_D$ is the $D$-dimensional identity matrix and $\beta > 0$ is a so-called temperature parameter, which typically is a hyper parameter used to control how random the output of a neural networks is. Here we will use it to distinguish between generalized and standard Bayesian setting. Further note that the prior variance of the connection weights $\sigma^2_0\in\mathbb{R}$ will play an important role in subsequent analysis. Finally, we write the normalized posterior as $\pi_\beta = \tilde\pi_{\beta}/Z_{\beta}$, with the normalizing constant $Z_\beta$ denoted as the marginal likelihood or the evidence.

In the case of a regression problem with Gaussian noise, the following statistical model is often assumed to relate the model $f$ to the data $y$;
\begin{align}
y_n = f(x_n;\theta) + \epsilon_n,\quad \epsilon_n \sim\mathcal{N}\left(0, \sigma^2I_M\right).
\end{align}
\noindent
Here, $\sigma^2$ is the noise variance, which is assumed known, or estimated from residuals. Under this model, the standard Bayesian posterior is recovered by setting $\beta=N\sigma^{-2}$ in \eqref{eq:gen_post}, which is equivalent to replacing $\mathcal{L}_N$ with a summed loss. In contrast, the generalized posterior based on the mean squared loss is obtained using $\beta=\sigma^{-2}$. For classification with softmax cross-entropy, there is no analogous Gaussian noise variance. Similarly,  $\beta=N$ corresponds to the summed categorical negative log-likelihood of the standard Bayesian posterior, whereas the generalized posterior based on the mean cross-entropy loss requires $\beta=1$.

\subsubsection{Laplace approximation of the generalized posterior\label{sec:gen_laplace}}

We define $l_\beta(\theta)$ as the negative log of \eqref{eq:gen_post}, i.e.;
\begin{align}
l_\beta(\theta) := -\log \tilde\pi_\beta = \beta\mathcal{L}_N - \log p(\theta) \propto \beta\mathcal{L}_N + \frac{1}{2\sigma^2_0}\lVert\theta\rVert^2_2.
\end{align}
\noindent
The most-likely {\it maximum a-posteriori} (MAP) parameter value solves 
\begin{align}
    \hat\theta = \argmin_{\theta} l_\beta({\theta}).
\end{align}
\noindent
We will use the pretrained weights $\theta_0$ in place of the MAP value, which is a common approximation (also used by e.g.\ \cite{ritter2018} and \cite{immer2021improving}) unless the network was trained using L2 weight decay with regularization strength consistent with the values of $1/(2\sigma^2_0)$ and $\beta$. In our approach, $\theta_0$ serves as the anchor point for the Laplace approximation, originally applied by \cite{mackay1992practical} to small neural networks. This approximation models the posterior distribution $\pi_\beta(\theta\mid y)$ as a Gaussian distribution, with a covariance structure defined from (the inverse of) the curvature of the log-likelihood function. The underpinning idea here is that the log likelihood quickly drops away (i.e.\ has high curvature) when the data is able to inform the parameters well. For generalized Bayes, the covariance structure is derived from the curvature of the loss function instead. To arrive at such an expression a second-order Taylor expansion of $l_\beta({\theta})$ around $\theta_0$ is made
\begin{align}
   l_\beta(\theta) \approx l_\beta(\theta_0) + \frac{1}{2}\left(\theta - \theta_0\right)^T\nabla^2_{\theta}l_\beta(\theta_0)\left(\theta - \theta_0\right),
    \label{eq:taylor2}
\end{align}
\noindent
where under the assumption that $\theta_0$ is sufficiently close to a stationary point of $l_\beta$, the first-order term drops out: $\nabla_{\theta}l_\beta(\theta_0)=0$. Given \eqref{eq:taylor2}, the posterior $\pi_\beta$ now becomes a Gaussian $\mathcal{N}(\theta_0, \Sigma)$:
\begin{align}
    \pi_\beta(\theta\mid y)\propto \exp\left[-l_\beta(\theta)\right]\propto\exp\left[-\frac{1}{2}\left(\theta - \theta_0\right)^T\Sigma^{-1}\left(\theta - \theta_0\right)\right],
\end{align}
\noindent
with 
\begin{align}
\Sigma = \left(\nabla^2_{\theta}l_\beta(\theta_0)\right)^{-1}
= \left(\beta\nabla^2_{\theta}\mathcal{L}_N(\theta_0) - \nabla^2_{\theta}\,\;\log p(\theta)\right)^{-1} = \left(\beta H_L + \frac{1}{\sigma_0^2}I_D\right)^{-1}.
\label{eq:Sigma}
 \end{align}
\noindent
Note that here $H_{L}$ is defined to be the Hessian of the mean loss, $H_L:=\nabla^2_{\theta}\mathcal{L}_N(\theta_0
)\in\mathbb{R}^{D\times D}$. 

The Gaussian form is thus a consequence of truncating the Taylor expansion around $\theta_0$ at the quadratic term, it is not statement that the full neural-network posterior is indeed normally distributed, which may well be multi modal and non-Gaussian. The Laplace approximation is therefore a pragmatic, local posterior estimate around one trained solution, not as a faithful global description of all plausible weights. We examine its usefulness through its posterior predictive behaviour, i.e.\ by the resulting posterior distribution of the neural-network output. 

The Laplace approximation gives a closed-form expression for the posterior on $\theta$. Unlike UQ approaches based on (low-rank) variational inference (e.g.\ \cite{jantre2024learning,Nafiz_Abeer_2024}), no additional training is therefore required. Still, computing and inverting $H_L$ is computationally intractable in neural networks due to the size of $\theta$. To overcome this limitation we will construct a computationally tractable closed-form expression for the posterior variance.

\subsubsection{Low-rank posterior variance}

By examining \eqref{eq:Sigma}, we observe that in the principal coordinate system spanned by the eigenvectors from $H_L p_j = \lambda_j p_j$, the posterior variances $\sigma_j^2$ become;
\begin{align}
    \sigma_j^2 = \frac{1}{\beta\lambda_j + \sigma_0^{-2}},\quad j=1\cdots,D.
    \label{eq:post_var}
\end{align}
\noindent
In practice, a low-rank framework is obtained by retaining only the first $d \ll D$ dominant eigenvalues $\lambda_j$ of an approximation to the Hessian. The quality of this approximation depends on how well the approximate Hessian captures directions of significant curvature (see Section \ref{sec:approx_Hessians}), as well as on the choice of prior variance $\sigma_0^2$.

It is not uncommon to find references which discuss Hessian approximations in the context of the sum loss, see e.g.\ \cite{kunstner2019limitations}. As discussed, in the present work we use the loss from generalized Bayes rather than the summed negative log-likelihood. This corresponds to a temperature-scaled posterior where the parameter $\beta$ controls how strongly the data loss is weighted relative to the prior (see \eqref{eq:gen_post}). For $\beta=\sigma^{-2}$ or $\beta=1$ (mean regression / classification loss), the likelihood contribution is normalized by the number of data points. This normalization prevents the curvature eigenvalues from scaling linearly with $N$, such that the curvature is governed by the shape of the mean loss landscape, not by the size of the dataset. This is important in our low-rank subspace setting, since as $N$ increases, the posterior variances \eqref{eq:post_var} decrease in the standard Bayesian setting with $\beta = N\sigma^{-2}$ or $\beta=N$. While in this setting posterior contraction is expected as $N$ grows, in a low-rank projected posterior this can become too aggressive, and the leading active directions become extremely narrow. Using the mean loss avoids this purely sample-size-driven contraction. It makes the eigenvalues describe the average local sensitivity of the trained model rather than the total accumulated data-fit curvature.

Regardless of the choice of $\beta$, the form $\sigma_j^2 = (\beta\lambda_j + 1/\sigma_0^2)^{-1}$ highlights the interplay between data and prior. Directions with large average curvature (large $\lambda_j$) lead to stronger posterior contraction, while in flat directions ($\lambda_j \to 0$) the posterior variance reverts to the prior, $\sigma_j^2 \to \sigma_0^2$. Thus, the prior variance acts as a baseline level of uncertainty in directions not informed by the data.

Since $\sigma_0^2$ appears explicitly in the posterior variance and controls the overall scale of uncertainty, it must be chosen carefully. We estimate $\sigma_0^2$ using an empirical Bayes procedure in Section \ref{sec:emp_bayes}. First, we discuss common choices for the approximate Hessian, and means to efficiently extract their $d$ dominant eigenpairs.

\subsubsection{Approximate Hessians\label{sec:approx_Hessians}}

In general, an issue with using the true Hessian in the covariance expression is that it is not guaranteed to be positive semi-definite. Several approximations are therefore commonly employed, most notably the generalized Gauss–Newton matrix $G$, which coincides with the true Hessian only under specific assumptions \cite{kunstner2019limitations}. Understanding the precise conditions under which these matrices agree, and how they differ otherwise, is essential for justifying their use as curvature surrogates. In the following, we briefly make these relationships explicit and discuss their implications for subspace-based uncertainty quantification.

One can decompose the Hessian into a term involving second derivatives of the model output, and a (more tractable) term that depends only on first-order derivatives. A natural approach is to apply the chain rule to \eqref{eq:loss}, differentiating with respect to $f$ before taking the $\theta$ derivative. The resulting gradient will have entries
\begin{align}
\left(\nabla_\theta \mathcal{L}_N\right)_i = \frac{1}{N}\sum_{n=1}^N\sum_{m=1}^M r_n^{(m)}\frac{\partial f_n^{(m)}}{\partial \theta_i},\quad i=1,\cdots,D,
\label{eq:grad_LN}
\end{align} 
where $r_n^{(m)}$ are residuals, defined as:
\begin{align}
 r_n^{(m)} :=f_n^{(m)} - y_n^{(m)}\quad\mathrm{or}\quad r_n^{(m)}:=p_n^{(m)} - y^{(m)}_n,
\end{align}
\noindent
for regression or classification. Note that for the latter, $r_n^{(m)}$ is a probability residual, which unlike its regression counterpart, is a function of all $M$ $f_n^{(m)}$ entries (see \eqref{eq:softmax}). Applying the chain rule to \eqref{eq:grad_LN} once more gives the Hessian entries:
\begin{align}
\left(\nabla^2_\theta\mathcal{L}_N\right)_{i,j} = \frac{1}{N}\sum_n\sum_m \left(r_n^{(m)}\frac{\partial^2f_n^{(m)}}{\partial\theta_i\theta_j} + 
\sum_k\frac{\partial r_n^{(m)}}{\partial f_n^{(k)}}\frac{\partial f^{(k)}_n}{\partial\theta_i}\frac{\partial f^{(m)}_n}{\partial\theta_j}\right),\quad i,j=1,\cdots,D.
\end{align}
\noindent
When written in matrix form we obtain:
\begin{align}
H_L(\theta):=\nabla^2_\theta\mathcal{L}_N = \frac{1}{N}\sum_n\sum_m r_n^{(m)}\nabla^2_\theta f_n^{(m)} + \frac{1}{N}\sum_n \left(\nabla_\theta f_n\right)^T H_{l_n} \left(\nabla_\theta f_n\right).
\label{eq:H}
\end{align}
\noindent
Here, $\nabla_\theta f_n\in\mathbb{R}^{M\times D}$ is the Jacobian of the output, and $H_{l_n}:=\nabla^2_f l_n\in\mathbb{R}^{M\times M}$ is a Hessian with $\left(H_{l_n}\right)_{m,k}=\partial^2l_n/\partial f_n^{(m)}\partial f^{(k)}_n = \partial r_n^{(m)}/\partial f_n^{(k)}$, leading to
\begin{align}
    H_{l_n} = I_M\quad\mathrm{and}\quad H_{l_n}=\mathrm{diag}(p_n) - p_np_n^T
    \label{eq:H_l_n}
\end{align}
\noindent
for regression and classification respectively.

The generalized Gauss–Newton (GGN) matrix $G$ is an approximation to $H_L$ that captures (partial) curvature information while avoiding second-order derivatives of the model, and is obtained by simply ignoring the first term of \eqref{eq:H};
\begin{align}
H_L(\theta)\approx G(\theta):= \frac{1}{N}\sum_n \left(\nabla_\theta f_n\right)^T H_{l_n} \left(\nabla_\theta f_n\right).
\label{eq:G}
\end{align}
\noindent
This is convenient as it drops the (expensive) Hessians of $f_n^{(m)}$. Moreover, from \eqref{eq:H} we can see that this is a well-justified approximation as argued by \cite{kunstner2019limitations}, in the sense that the neglected term vanishes when the residuals are small. For \eqref{eq:G} to be exact, it would be sufficient if all residuals are zero. However, this is unrealistic, and would be indicative of overfitting. The authors of \cite{immer2021improving} therefore prefer an alternative justification for the use of $G$, namely that it is equivalent to replacing the network by its first-order Taylor expansion $f_{lin}$ around $\theta_0$, for which $\nabla^2_\theta f_{lin}(\theta_0)=0$ holds. This leads to so-called linearized Laplace predictive distributions, discussed in Section \ref{sec:pred}.

Finally, we note that empirical work has shown that the loss gradient $\nabla_\theta\mathcal{L}_N$ aligns strongly with the leading eigenvectors of the Hessian, see \cite{gur2018gradient}, implying that optimization proceeds primarily within a subspace spanned by a small number of dominant curvature directions. This observation provides support for low-rank posterior approximations involving (approximate) Hessians; if training mainly explores a restricted subspace, then uncertainty quantification should focus on these same directions. Our approach leverages this structure by constructing a posterior covariance aligned with the dominant eigenspace of $G(\theta)$. As mentioned, we use the term active subspace to denote the dominant eigenspace of the generalized Gauss-Newton curvature of the empirical mean loss, which is a slight departure from its original definition \cite{constantine2014active}. That said, the overarching idea is similar, as is the linear projection to the active subspace;
\begin{align}
w_0 = P^T\theta_0 \in\mathbb{R}^d,
\end{align}
\noindent
here applied to the connection weights. $P\in\mathbb{R}^{D\times d}$ contains the $d\ll D$ dominant (orthonormal) eigenvectors of $G$. In order to form $P$ for large neural networks, we first discuss matrix-free eigenvalue methods.

\subsubsection{Matrix-free computation of dominant eigenpairs\label{sec:matrix_free}}

In practice, computing the dominant eigenvalues / vectors of $G\in\mathbb{R}^{D\times D}$ will be complicated by the dimension of $G$. For large enough $D$, even storing $G$ in memory will not be possible. However, we can use the Lanczos algorithm \cite{chen2026lanczos, lanczos1950iteration} to approximate the leading eigenpairs, which only requires access to matrix-vector products $Gv$. These products can be computed using automatic differentiation without ever forming $G$. If we write $J_n:=\nabla_\theta f_n\in\mathbb{R}^{M\times D}$ for brevity, we obtain
\begin{align}
Gv = \frac{1}{N}\sum_n J_n^T H_{l_n} J_nv,
\label{eq:Gv}
\end{align}
\noindent
see \eqref{eq:G}. In general, by the chain rule, the Jacobian-vector product $J_nv\in\mathbb{R}^{M\times 1}$ equals
\begin{align*}
J_nv = \left.\frac{\D}{\D\epsilon} f(x_n; \theta + \epsilon v)\right|_{\epsilon=0},
\label{eq:pearlmutter_trick}
\end{align*}
\noindent
which in turn can be computed with forward-mode automatic differentiation, see e.g.\ \cite{pearlmutter1994fast}. 
%
Once $J_nv$ is available at the output, pre-multiplication with the known $H_{l_n}$ \eqref{eq:H_l_n} yields $u_n:=H_{l_n}J_nv\in\mathbb{R}^{M\times 1}$. If we define the auxiliary scalar loss function $\tilde{L}:=u_n^Tf$ such that its derivative equals $\D\tilde{L}/\D f = u_n$, then once again by the chain rule;
\begin{align*}
\frac{\D\tilde{L}}{\D\theta} = J_n^Tu_n = J_n^TH_{l_n}J_n v\in\mathbb{R}^{D},
\end{align*}
\noindent
which equals the $n$-th term of \eqref{eq:Gv} and can be computed using standard back propagation. Hence, each of the $N$ contributions to $Gv$ can be evaluated using one forward- and one backward-pass through the network, without ever explicitly forming the Jacobian $J_n$ or the matrix $G$. In practice one uses mini-batches in place of a term-by-term computation. In addition, libraries such as PyTorch have dedicated subroutines for computing the required Jacobian-vector products.

Access to $Gv$ enables scalable computation of the dominant eigenspace of G in high-dimensional settings. Briefly, we use the Lanczos algorithm, a well-known iterative method for approximating a few extremal eigenvalues and eigenvectors of a symmetric matrix $G \in \mathbb{R}^{D \times D}$ using only matrix--vector products. Starting from an initial unit vector $v_1$, it constructs an orthonormal basis $\{v_1, \dots, v_l\}$ of the Krylov subspace $\mathrm{span}\{v_1, Gv_1, \dots, G^{l-1}v_1\}$ via the recursion
\begin{align}
v_{k+1} = \frac{G v_k - b_{k-1} v_{k-1} - a_k v_{k}}{b_k}, \quad b_k = \lVert G v_k - b_{k-1} v_{k-1} - a_k v_{k} \rVert_2, \quad a_k = v_k^TGv_k.
\end{align}
with $b_0 = 0$, which follows from a Gram-Schmidt procedure. This yields a low-dimensional symmetric tridiagonal matrix $T_l \in \mathbb{R}^{l \times l}$ with $d < l \ll D$ constructed from $a_k$, $b_k$ coefficients (see e.g.\ \cite{chen2026lanczos}) for which
\begin{align}
T_l = V_l^T G V_l
\end{align}
holds, where $V_l = [v_1, \dots, v_l]\in\mathbb{R}^{D\times l}$. The eigenvalues of $T_l$ directly approximate the leading eigenvalues of $G$, and $V_lU_l\in\mathbb{R}^{D\times l}$ approximate the corresponding eigenvectors, $U_l\in\mathbb{R}^{l\times l}$ being the eigenvectors of $T_l$. 

The first $d$ columns of $V_lU_l$ define our projection matrix $P\in\mathbb{R}^{D\times d}$. Note that in order to bring the error down as in Figure \ref{fig:eigvals_p1}, it is important to select a sufficiently high number of basis functions $l>d$ to span the Krylov subspace. We used $l=d+10$. Crucially, note that the method requires only repeated evaluations of $Gv$, making it well-suited for large-scale settings where $G$ cannot be formed explicitly.

\subsubsection{Empirical generalized Bayesian subspace inference for the prior variance\label{sec:emp_bayes}}

As previously mentioned, the value of the prior variance plays an important role in \eqref{eq:post_var}, and must be chosen with care. However, defining a plausible prior on the connection weights $\theta$ is challenging due to their lack of physical meaning. Empirical Bayesian inference \cite{carlin2000empirical} provides a solution in this context by allowing us to estimate $\sigma_0^2$ directly from the data $y$. That said, the severe overparameterization of neural networks again poses specific challenges. We found that applying empirical Bayes directly in the full, high-dimensional parameter space forces the optimization toward excessively high $\sigma _{0}$ values to compensate for the many $D-d$ flat, uninformative directions. To resolve this, we perform empirical Bayes in the projected parameter space defined by the active subspace $w=P^T\theta\in\mathbb{R}^d$. Again, let $P\in\mathbb R^{D\times d}$ contain the dominant eigenvectors of $G$ and define
\begin{align}
w_0=P^T\theta_0,
\qquad
\theta(w)=\theta_0+P(w-w_0).
\end{align}
\noindent
The loss restricted to the active subspace is then
\begin{align*}
\mathcal L_N(w,y):=\mathcal L_N(\theta(w),y)
\end{align*}
\noindent
Our analysis will proceed similar to Section \ref{sec:gen_laplace}. Let $\tilde\pi_{\beta, w}$ be the generalized unnormalized posterior in the $w$ subspace, i.e.;
\begin{align}
\tilde\pi_{\beta,w}\left(w\mid y,\sigma^2_0\right) = \exp\left[-\beta\mathcal{L}_N(w, y)\right]p(w\mid\sigma^2_0),
\end{align}
\noindent
with the prior $p(w\mid\sigma^2_0) = \mathcal{N}\left(0, \sigma^2_0I_d\right)$. We find $\sigma^2_0$ by maximizing the generalized evidence:
\begin{align}
Z_{\beta,w}(y \mid \sigma_0^2) = \int \tilde\pi_{\beta,w}\left(w\mid y,\sigma^2_0\right) \D w.
\end{align}
\noindent
The log density is given by
\begin{align}
\log Z_{\beta,w}(y \mid \sigma_0^2) = \log \int \exp\left[ -l_w(w\mid\sigma^2_0)\right]\;\D w.
\label{eq:log_Z}
\end{align}
\noindent
with $l_w(w\mid\sigma^2_0)$ defined as the negative log unnormalized generalized posterior density:
\begin{align}
l_w:=-\log\tilde\pi_{\beta,w}(w\mid\sigma^2_0) =  \beta\mathcal{L}_N(w, y) - \log p(w\mid\sigma^2_0)
\end{align}
\noindent
We now insert a Taylor expansion for $l_w$ around $w_0=P^T\theta_0$ into \eqref{eq:log_Z}, to give rise to a Laplace approximation;
\begin{align}
\log Z_{\beta, w}(y \mid \sigma_0^2) \approx -l_w(w_0\mid\sigma^2_0) + \log\int \exp\left[-\frac{1}{2}\left(w - w_0\right)^T[\nabla^2_{w}l_w(w_0)]\left(w - w_0\right)\right] \;\D w.
\label{eq:log_marginal_ll0}
\end{align}
\noindent
As in the full-space Laplace approximation, we assume that $w_0$ is sufficiently close to a stationary point of $l_w$, so that the first-order term in the Taylor expansion can be neglected. The second log term contains the integral of the unnormalized Laplace posterior in the active subspace, with $\nabla^2_{w}l_w= P^T\;\nabla^2_{\theta}l_\beta\, P = P^T\left[\beta\nabla^2_{\theta} \mathcal{L}_N - \nabla^2_{\
\theta}\log p(\theta_0\mid\sigma^2_0)\right]P = \beta P^TH_LP + \sigma^{-2}_0 I_d$, and therefore equals the reciprocal of the Gaussian normalizing constant. Defining $H_w:=P^TH_LP$, \eqref{eq:log_marginal_ll0} then becomes;
\begin{align}
\log Z_{\beta,w}(y \mid \sigma_0^2) & \propto -l_w(w_0\mid\sigma^2_0) - \frac{1}{2}\log \left(\det\left[\beta H_w + \frac{1}{\sigma_0^2}I_d\right]\right) \nonumber\\
&\propto -\beta\mathcal{L}_N(w_0, y) + \log p(w_0\mid\sigma^2_0) - \frac{1}{2}\log \left(\det\left[\beta H_w + \frac{1}{\sigma_0^2}I_d\right]\right)  \nonumber\\
&\propto \log p(w_0\mid\sigma^2_0) - \frac{1}{2}\log \left(\det\left[\beta H_w + \frac{1}{\sigma_0^2}I_d\right]\right) \nonumber\\
&\propto -\frac{1}{2\sigma^2_0}\lVert w_0\rVert^2_2 - d\log\sigma_0 - \frac{1}{2}\sum_{i=1}^d \log\left(\beta\lambda_i + \frac{1}{\sigma^2_0}\right),
\end{align}
\noindent
The loss $\mathcal{L}_N$ does not depend explicitly on $\sigma_0^2$, and is therefore considered constant in this approximation. In the last line we swap out $H_L$ with an approximate Hessian discussed in Section \ref{sec:approx_Hessians} (i.e.\ $H_w\approx P^TGP$), and the low-rank structure of $H_w$ allows us to then compute the determinant using only the $d$ dominant eigenvalues obtained from the matrix-free method of Section \ref{sec:matrix_free}. Setting $\alpha = \sigma^{-2}_0$, we now find that
\begin{align}
0 &= \frac{\D}{\D\alpha} \left(-\frac{\alpha}{2}\lVert w_0\rVert^2_2  + \frac{d}{2}\log\alpha - \frac{1}{2}\sum_{i=1}^d \log\left(\beta\lambda_i + \alpha\right)\right) \nonumber\\
 &= -\frac{1}{2}\lVert w_0\rVert^2_2 + \frac{d}{2\alpha} - \frac{1}{2}\sum_{i=1}^d\frac{1}{\beta\lambda_i + \alpha}.
\end{align}
\noindent
Solving for the projected weight norm;
\begin{align}
\lVert w_0\rVert^2_2 = \sum_{i=1}^d \frac{1}{\alpha} - \frac{1}{\beta\lambda_i + \alpha} = \sum_{i=1}^d \sigma^2_0 - \sigma^2_i,
\label{eq:sigma_0_im}
\end{align}
\noindent
with $\sigma^2_i$ being the posterior variance \eqref{eq:post_var}. Therefore, the resulting $\sigma_0^2$ sets the prior scale such that the magnitude of the projected parameter vector is consistent with the amount of variance reduction in the active subspace.

Note that performing this analysis in the full parameter space will lead to an expression involving $\lVert\theta_0\rVert^2_2$ in place of $\lVert w_0\rVert^2_2$, which will be very large for overparameterized neural networks. This then leads to an excessively large prior variance $\sigma^2_0$.

Although \eqref{eq:sigma_0_im} defines an implicit and non-linear equation for $\sigma_0^2$, the corresponding function is smooth and strictly decreasing in $\alpha$, and therefore admits a unique solution. This solution can be efficiently obtained using standard one-dimensional root-finding methods (we use the bisection method).

\subsubsection{Calibrating d\label{sec:d}}

The dimension of the active subspace $d$ is a hyperparameter that must be set. For regression problems, we will do so by evaluating the coverage of the output confidence intervals on a calibration data set $(x^*_n, y^*_n)$, $n=1,\cdots,N_{cal}$. Let $I_{\gamma}(x)$ be the $1-\gamma$ confidence interval for $\gamma\in [0,1]$. Our value of $d$ minimizes
\begin{align}
d^* = \argmin_d \frac{1}{N_{cal}} \sum_{n=1}^{N_{cal}} \mathbbm{1} \left[ y^*_n\in I_\gamma(x^*_n)\right] \geq 1-\gamma.
\end{align}
\noindent
Here $\mathbbm{1}[A]$ is an indicator function returning 1/0 if $A$ is true/false. Note that for each $d$, we must perform the empirical Bayesian analysis of Section \ref{sec:emp_bayes}, such that $d$ and $\sigma_0$ are estimated jointly. Calibration in the context of classification entails making sure the correct class in contained in the $1-\gamma$ set with (approximately) probability $1-\gamma$.

A similar calibration procedure can be applied to a classification network \cite{yang2024bayesian, guo2017calibration}. However, if a classification network is already well calibrated, the additional stochasticity introduced via weight perturbations may diminish calibration, see the Discussion section.

\subsubsection{Posterior prediction\label{sec:pred}}

The posterior variance \eqref{eq:post_var} corresponds to the Laplace approximation restricted to the dominant eigenspace of the curvature matrix. In the full parameter space, directions orthogonal to this subspace have variance $\sigma_0^2$, corresponding to the prior. While these directions are individually flat, they are vast in number ($D - d \gg d$), and can collectively contribute significantly to predictive uncertainty, often resulting in overly wide confidence intervals when posterior samples are propagated through the non-linear network. 

Similar to e.g.\ \cite{jantre2024learning, faller2026low}, we therefore omit these directions and obtain a low-rank approximation that concentrates uncertainty in the data-informed subspace. Sampling from this approximate posterior is then performed via
\begin{align}
\theta = \theta_0 + Pz,\quad z\sim\mathcal{N}\left(0, \mathrm{diag}(\sigma^2_j)\right), \quad \sigma^2_j=\frac{1}{\beta\lambda_j + \sigma^{-2}_0},\quad j=1,\cdots,d.
\end{align}
\noindent
Using these samples during a forward pass we can approximate expectations such as
\begin{align}
f(x\mid y) = \int f(x;\theta) \pi_\beta(\theta\mid y)\D\theta,
\end{align}
via Monte Carlo sampling, pushing the Laplace posterior through the network. As the Laplace posterior is an approximation, the non linear nature of the network may push the mean prediction away from the MAP prediction. 

This underfitting problem of sampled Laplace does not primarily mean that the Gaussian posterior family is insufficiently expressive. Rather, it refers to a degradation of the posterior predictive mean caused by propagating local Gaussian weight samples through the non-linear network. As mention in Section \ref{sec:approx_Hessians}, linearized Laplace \cite{immer2021improving} avoids this by making the a local linear approximation in the predictive model, thereby keeping the posterior predictive mean fixed at the MAP prediction:
\begin{align}
    f(x;\theta) \approx f_{lin}(x; \theta) = f(x;\theta_0) + J(x)\left(\theta - \theta_0\right),
    \label{eq:f_lin}
\end{align}
\noindent
where $J(x):=\nabla_\theta f(x;\theta_0)\in\mathbb{R}^{M\times D}$. The combination of the linear output and the Gaussian distribution of $\theta$ also leads a closed-form predictive covariance $\Sigma_f=J\Sigma J^T$, where $\Sigma=P\Lambda P^T$, with $\Lambda:=\mathrm{diag}(\sigma^2_j)$ is our low-rank Laplace posterior covariance. In the case of scalar regression the (pointwise) posterior predictive variance becomes
\begin{align}
    \sigma^2_f(x) = \sum_{j=1}^d \sigma^2_j \left(J(x) p_j\right)^2\in\mathbb{R},
    \label{eq:sigma_f}
\end{align}
with $p_j\in\mathbb{R}^D$ denoting the j-th column vector of $P$.

\backmatter

\bmhead{Supplementary information}

For this article no Supplementary Materials are available. 

\bmhead{Acknowledgements}
We thank Dr Ketan Maheshwari at the US Department of Energy Oak Ridge National Laboratory and Dr Alessandro Tibo at AstraZeneca for helpful discussions and support with training REINVENT. We also thank Dr Laura Harbach at Brunel University London for initiating the REINVENT UQ study.



\section*{Declarations}


\subsection*{Funding}
P.V.C. is grateful for funding from the UK Engineering and Physical
Sciences Research Council under the following project “UK Consortium on Mesoscale
Engineering Sciences (UKCOMES)” (Grant No. EP/R029598/1).

\subsection*{Competing interests}
The authors declare no competing interests.

\subsection*{Ethics approval and consent to participate}
Not applicable.

\subsection*{Consent for publication}
Not applicable.

\subsection*{Data and code availability}
The data used to train the neural networks, a pre-trained REINVENT model and Jupyter notebooks to reproduce the results of this article are available at \cite{edeling2026glas}.

\subsection*{Materials availability}
Not applicable.

\subsection*{Author contribution}

W.N.E. developed the methodology, implemented the algorithms, performed the numerical experiments, analysed the results, and wrote the initial manuscript. P.V.C. contributed to the conceptual framing, interpretation of results, funding acquisition and manuscript revision.  Both authors reviewed and approved the final manuscript.

\subsection*{Declaration on the use of artificial intelligence tools}

The authors used OpenAI's ChatGPT 5.5 as a tool for language editing. The tool was also used to discuss manuscript presentation, related-work positioning, and the interpretation of selected numerical results. All scientific claims, mathematical derivations, algorithms, references, numerical experiments, and conclusions were reviewed, verified, and finalized by the authors. The authors assume full responsibility for the content of the manuscript.






\pagebreak
\begin{appendices}






\section{ChEMBL vocabulary \label{app:voc}}

Here we describe the ChEMBL vocabulary in Table \ref{tab:smiles_vocabulary} and display some example chemical compounds in Table \ref{tab:smiles}.

\begin{table*}[ht]
\centering
\caption{Vocabulary used by the autoregressive SMILES generator. The vocabulary contains start and stop tokens, bond symbols, branch symbols, ring-closure indices, atom tokens, and charged or aromatic atom tokens.}
\label{tab:smiles_vocabulary}
\small
\begin{tabular}{cll|cll}
\hline
Index & Token & Description & Index & Token & Description \\
\hline\hline
0  & \texttt{\$}       & Stop/end token
& 17 & \texttt{Br}      & Bromine atom \\

1  & \texttt{\string^} & Start token
& 18 & \texttt{C}       & Aliphatic carbon atom \\

2  & \texttt{\#}       & Triple bond
& 19 & \texttt{Cl}      & Chlorine atom \\

3  & \texttt{\%10}     & Two-digit ring closure
& 20 & \texttt{F}       & Fluorine atom \\

4  & \texttt{(}        & Branch opening
& 21 & \texttt{N}       & Aliphatic nitrogen atom \\

5  & \texttt{)}        & Branch closing
& 22 & \texttt{O}       & Oxygen atom \\

6  & \texttt{-}        & Single bond
& 23 & \texttt{S}       & Sulfur atom \\

7  & \texttt{1}        & Ring closure index
& 24 & \texttt{[N+]}    & Positively charged nitrogen \\

8  & \texttt{2}        & Ring closure index
& 25 & \texttt{[N-]}    & Negatively charged nitrogen \\

9  & \texttt{3}        & Ring closure index
& 26 & \texttt{[O-]}    & Negatively charged oxygen \\

10 & \texttt{4}        & Ring closure index
& 27 & \texttt{[S+]}    & Positively charged sulfur \\

11 & \texttt{5}        & Ring closure index
& 28 & \texttt{[n+]}    & Positively charged aromatic nitrogen \\

12 & \texttt{6}        & Ring closure index
& 29 & \texttt{[nH]}    & Protonated aromatic nitrogen \\

13 & \texttt{7}        & Ring closure index
& 30 & \texttt{c}       & Aromatic carbon atom \\

14 & \texttt{8}        & Ring closure index
& 31 & \texttt{n}       & Aromatic nitrogen atom \\

15 & \texttt{9}        & Ring closure index
& 32 & \texttt{o}       & Aromatic oxygen atom \\

16 & \texttt{=}        & Double bond
& 33 & \texttt{s}       & Aromatic sulfur atom \\
\hline\hline
\end{tabular}
\end{table*}

\begin{table}[ht]
\centering
\caption{10 SMILES strings from the ChEMBL database.}
\label{tab:smiles}
\begin{tabular}{cl}
\hline
No. & SMILES \\
\hline\hline
1  & \texttt{CC(C)C(NC(=O)C(NC(=O)C1OC2OC(C)(C)OC2C2OC(C)(C)OC12)C(C)C)C(O)=O} \\
2  & \texttt{Fc1ccc(cc1)C(=O)NN1C(SCc2ccc(cc2)N(=O)=O)=Nc2ccccc2C1=O} \\
3  & \texttt{CC(C)CN1Cc2cnnn2-c2ccc(cc2C1)-c1ccccc1} \\
4  & \texttt{CCN(C(=O)CSc1nnnn1-c1ccccc1)C1=C(N)N(Cc2ccccc2)C(=O)NC1=O} \\
5  & \texttt{CC(C)N1CCN(CC1)C(=O)N1c2ccccc2Sc2ccccc12} \\
6  & \texttt{CC1(C)Cc2cc(OCc3ccc(cc3)-c3nn[nH]n3)c(Cl)c(Cl)c2C1=O} \\
7  & \texttt{Cc1cccc(CSc2ccc(cn2)C(=O)Nc2ccc(F)cc2)c1F} \\
8  & \texttt{CCC(N1N=C(O)C2=Nc3cc(Cl)ccc3C(=O)C2=C1O)c1ccccn1} \\
9 & \texttt{CC(N1C(=O)C2OC3(C)C=C(C)C=NC3C2C1=O)c1ccccc1} \\
10 & \texttt{Cc1nc(cs1)-c1cccc(NC(=O)c2ccc3nc4C(=O)NCCCn4c3c2)c1} \\
\hline\hline
\end{tabular}
\end{table}

\end{appendices}


\bibliography{sn-bibliography}

\end{document}